\documentclass[10pt,twocolumn,letterpaper]{article}

\usepackage{cvpr}
\usepackage{times}
\usepackage{epsfig}
\usepackage{graphicx}
\usepackage{amsmath}
\usepackage{amssymb}
\usepackage{multirow}
\usepackage{color}
\usepackage{epstopdf}
\usepackage{cases}
\usepackage[scriptsize]{subfigure}
\usepackage{caption}
\usepackage{float}
\usepackage{lipsum}
\usepackage{authblk}
\usepackage{enumerate}
\usepackage{graphicx}

\makeatletter
\renewcommand{\@thesubfigure}{\hskip\subfiglabelskip}
\renewcommand{\tiny}{\fontsize{6.4pt}{\baselineskip}\selectfont}
\makeatother

\usepackage[breaklinks=true,bookmarks=false]{hyperref}

\cvprfinalcopy 

\def\cvprPaperID{****} 

\begin{document}

\title{Cascaded Partial Decoder for Fast and Accurate Salient Object Detection}


\author[1,2]{\rm Zhe Wu}
\author[1,2,3]{\rm Li Su}
\author[1,2,3]{\rm Qingming Huang}
\affil[1]{School of Computer Science and Technology, University of Chinese Academy of Sciences (UCAS), Beijing, China}
\affil[2]{Key Lab of Big Data Mining and Knowledge Management, UCAS, Beijing, China}
\affil[3]{Key Lab of Intelligent Information Processing, Institute of Computing Technology, Chinese Academy of Sciences, Beijing, China}
\affil[ ]{\textit {zhe.wu@vipl.ict.ac.cn, \{suli, qmhuang\}@usas.ac.cn}}

\maketitle
\thispagestyle{empty}
\pagestyle{empty}

\begin{abstract}
      Existing state-of-the-art salient object detection networks rely on aggregating multi-level features of pre-trained convolutional neural networks (CNNs). Compared to high-level features, low-level features contribute less to performance but cost more computations because of their larger spatial resolutions. In this paper, we propose a novel Cascaded Partial Decoder (CPD) framework for fast and accurate salient object detection. On the one hand, the framework constructs partial decoder which discards larger resolution features of shallower layers for acceleration. On the other hand, we observe that integrating features of deeper layers obtain relatively precise saliency map. Therefore we directly utilize generated saliency map to refine the features of backbone network. This strategy efficiently suppresses distractors in the features and significantly improves their representation ability. Experiments conducted on five benchmark datasets exhibit that the proposed model not only achieves state-of-the-art performance but also runs much faster than existing models. Besides, the proposed framework is further applied to improve existing multi-level feature aggregation models and significantly improve their efficiency and accuracy.
\end{abstract}

\section{Introduction}

Recently, deep learning has achieved surprising performance in salient object detection for it providing abundant and discriminative
image representations. The early deep saliency methods~\cite{2016ELD,2015MDF,2015LEGS} utilize CNNs to predict saliency scores of image regions and obtain accurate saliency maps with high computational complexity. In the following works, fully convolutional network (FCN)~\cite{2015FCN} based encoder-decoder architecture is widely applied for salient object detection. The encoder is the pre-trained image classification model (\emph{e.g.} VGG~\cite{2014VGG} and ResNet~\cite{2016ResNet}) which provides multi-level deep features: the high-level features with low resolutions represent semantic information, and the low-level features with high resolutions represent spatial details. In the decoder, these features are combined to generate accurate saliency maps. Researchers have developed various decoders~\cite{2018DSS,2016DCL,2016DHS,2018PiCANet,2017NLDF,2018BMPM,2017Amulet} to integrate low-level and high-level features.

\begin{figure}[t]
\centering
\includegraphics[width=\linewidth]{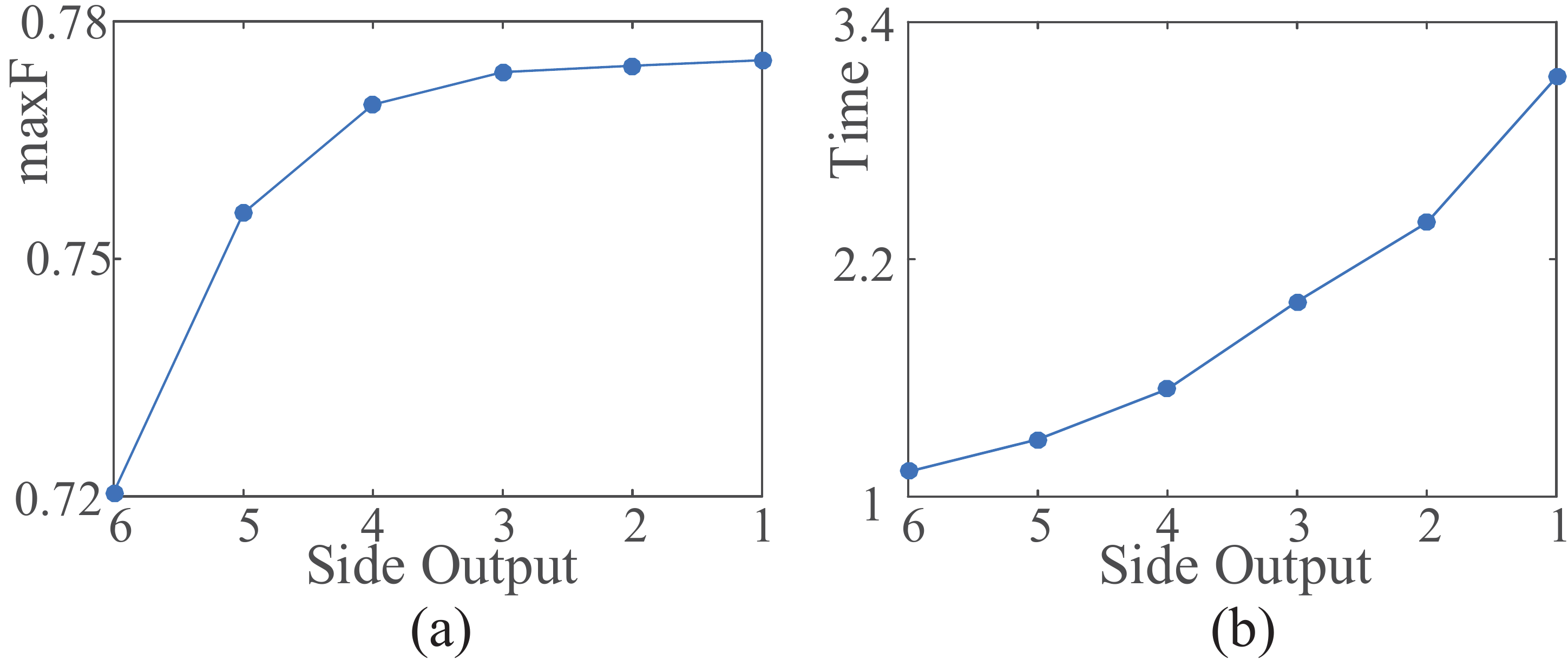}
\caption{\small \em (a) Maximum F-measure of six side outputs of the original DSS~\cite{2018DSS} model in PASCAL-S~\cite{Dataset-PASCAL-S} dataset. (b) We set inference time of backbone network as $1$, and show inference time of each side output here. The performance growth is getting slower and the inference time rapidly increases when gradually integrating features from high-level $6$ to low-level $1$.}
\label{DSS_SideOutputs}
\end{figure}

However, two drawbacks exist in these deep aggregation methods. On the one hand, compared to high-level features, low-level features contribute less to the performance of deep aggregation methods. In Fig.~\ref{DSS_SideOutputs}(a), we present performances of different side outputs of the DSS~\cite{2018DSS} model. It is obvious that the performance tends to saturate quickly when gradually aggregating features from high-level to low-level. On the other hand, due to the large resolutions of low-level features, integrating them with high-level features obviously enlarges the computational complexity as shown in Fig.~\ref{DSS_SideOutputs}(b). However, detecting and segmenting salient objects should be fast since this process is often a preprocessing stage to more complex operations~\cite{SalObjSurvey}. In consequence, it is essential to design a mechanism to eliminate the impact of low-level features on computational complexity while ensuring the performance.

\begin{figure}
  \centering
  \includegraphics[width=0.9\linewidth]{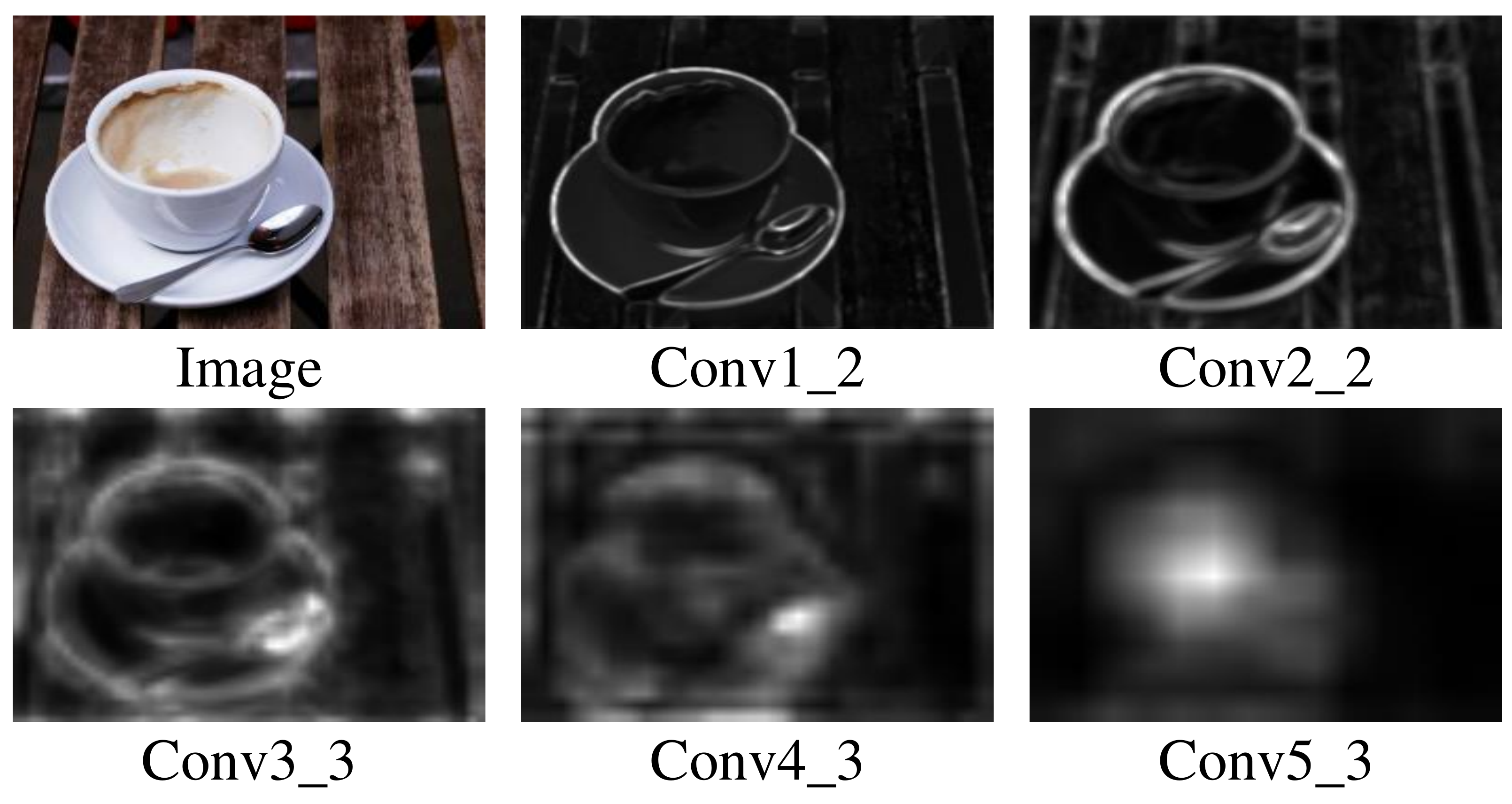}\\
  \caption{\emph{The original image and five-level feature maps from VGG$16$~\cite{2014VGG}. The Conv$3\_3$ feature still retains edge information. Hence the Conv$1\_2$ and Conv$2\_2$ features with large resolutions are not under consideration in the proposed framework.}}
  \label{VGGFeatures}
\end{figure}

When CNNs go deep, feature gradually changes from low-level representation to high-level representation. Hence deep aggregation models may recover spatial details of saliency maps when only integrating features of deeper layers. In Fig.~\ref{VGGFeatures}, we show examples of multi-level feature maps of VGG$16$~\cite{2014VGG}. Compared to low-level features of Conv$1\_2$ and Conv$2\_2$ layers, the feature of Conv$3\_3$ layer also reserve edge information. Besides, background regions in feature maps may result in inaccuracy of saliency maps. Previous works make use of adaptive attention mechanism~\cite{2018PiCANet,2018BMPM} to solve this problem. However, the effect of this mechanism relies on the accuracy of the attention map. Since fusing features of deeper layers will generate relatively precise saliency map, we can directly use this map to refine features.

In this paper, we propose a novel cascaded partial decoder framework, which discards features of shallower layers to ensure high computational efficiency and then refine features of deeper layers to improve their representation ability. We modify the standard straight backbone network to a bifurcated one. This new backbone network contains two branches with the same architecture. We construct partial decoder which only aggregates features in each branch. In order to further accelerate the model, we design a fast and efficient context module to abstract discriminative features and integrate them in an upsampling-concatenating way. Then we propose a cascaded optimization mechanism which utilizes initial saliency map of the first branch to refine features of the second branch. In order to uniformly segment the whole salient objects, we propose a holistic attention module to allow the initial saliency map cover more useful information. In addition, the proposed framework can be utilized to improve existing deep aggregation models. When embedding their decoders in our framework, the accuracy and efficiency will be significantly improved. Our contributions are summarized as follows:

\begin{enumerate}[(1)]
  \item We propose a novel cascaded partial decoder framework, which discards low-level features to reduce the complexity of deep aggregation models, and utilizes generated relatively precise attention map to refine high-level features to improve the performance.
  \item Experimental results on five benchmark datasets demonstrate that the proposed model not only achieves state-of-the-art performance but also runs much faster than existing models.
  \item Our framework can be applied to improve existing deep aggregation models. The efficiency and accuracy of improved models will both be significantly improved compared to the original models.
\end{enumerate}

\section{Related Work}
Over the past two decades, researchers have developed a large amount of saliency detection algorithms. Traditional models extract hand-crafted features and are based on various saliency assumptions~\cite{2009FT,2015GlobalContrast,1998Itti,2014wCtr}. More details about traditional methods are concluded in~\cite{SalObjSurvey,2015SalObjBenchmark}.
Here we mainly discuss deep learning based saliency detection models.

Early works utilize CNNs to determine whether image regions are salient or not~\cite{2016ELD,2015MDF,2015LEGS,2015MCDL}. Although these models have achieved much better performance than traditional methods, it is time-consuming to predict saliency scores for image regions. Then researchers develop more effective models based on the successful fully convolutional network~\cite{2015FCN}. Li~\emph{et al.}~\cite{2016MTDS} set up a unified framework for salient object detection and semantic segmentation to effectively learn the semantic properties of salient objects. Wang~\emph{et al.}~\cite{2016RFCN} leverage cascaded fully convolutional networks to continuously refine previous prediction maps.

\begin{figure*}[ht]
  \centering
  \includegraphics[width=\linewidth]{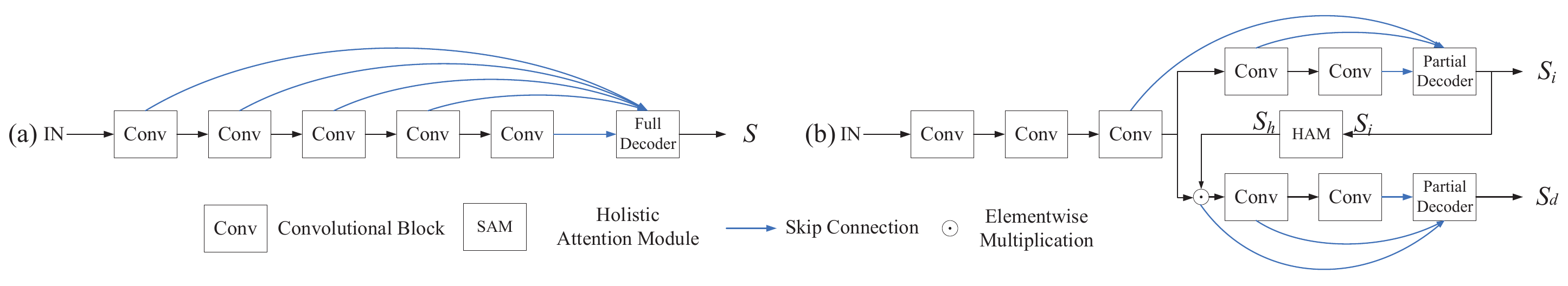}\\
  \caption{\emph{(a) Traditional encoder-decoder framework, (b) The proposed cascaded partial decoder framework. We use VGG$16$~\cite{2014VGG} as the backbone network. Traditional framework generates saliency map $S$ by adopting full decoder which integrates all level features. The proposed framework adopts partial decoder, which only integrates features of deeper layers, and generates an initial saliency map $S_{i}$ and the final saliency map $S_{d}$.}}\label{Structure}
\end{figure*}

Recently, researchers have proved that fusing multi-level features further improves the performance of dense prediction tasks~\cite{2015HyperColumn,2015UNet}. In CNNs, high-level features provide semantic information, and low-level features contains spatial details which are helpful for refining object boundaries. Many works~\cite{2018DSS,2016DCL,2016DHS,2018PiCANet,2017NLDF,2018BMPM,2017Amulet} follow this strategy and precisely segment salient objects. Li~\emph{et al.}~\cite{2016DCL} directly integrate multi-level features to obtain more advanced feature representation. Liu and Han~\cite{2016DHS} first make a coarse global prediction, and then hierarchically and progressively refine the details of saliency maps step by step via integrating local context information. Hou~\emph{et al.}~\cite{2018DSS} introduce short connections to the skip-layer structures within the HED~\cite{2015HED} architecture. Luo~\emph{et al.}~\cite{2017NLDF} segment salient objects by combining local contrast feature and global information through a multi-resolution $4\times 5$ grid network. Zhang~\emph{et al.}~\cite{2017Amulet} first integrate multi-level feature maps into multiple resolutions, which simultaneously incorporate semantic information and spatial details. Then this work predicts the saliency map in each resolution and fuses them to generate the final saliency map. In~\cite{2018BMPM}, the work extracts context-aware multi-level features and then utilizes a bi-directional gated structure to pass messages between them. Liu~\emph{et al.}~\cite{2018PiCANet} leverage global and local pixel-wise contextual attention network to capture global and local context information. Then these modules are incorporated with U-Net architecture to segment salient objects. In this paper, we argue that low-level features always contribute less than high-level features. However, they need more computation cost than high-level features owing to their larger spatial resolutions. Hence we propose a novel cascaded partial decoder framework for salient object detection, which does not consider low-level features and utilizes generated saliency map to refine high-level features.

\section{The Proposed Framework}
In this paper, we propose a novel cascaded partial decoder framework which contains two branches. In each branch, we design a fast and effective partial decoder. The first branch generates an initial saliency map which is utilized to refine the features of the second branch. Besides, we propose a holistic attention module to segment the whole objects uniformly.

\subsection{Mechanism of the Proposed Framework}\label{3.1}
We design the proposed model on the basis of VGG$16$ network, which is the most widely utilized backbone network in deep salient object detection models. For an input image I with size $H\times W$, we can abstract features at five levels, which are denoted as $\{f_{i}, i = 1,...,5\}$ with resolutions $[\frac{H}{2^{i-1}}, \frac{W}{2^{i-1}}]$. The decoders proposed in previous works~\cite{2018BMPM,2017Amulet}, which are called full decoder in this paper, integrate all features to generate saliency map $S$. A unified architecture of the full decoder is shown in Fig.~\ref{Structure}(a) and it can be represented by $D_{T} = g(f_{1}, f_{2}, f_{3}, f_{4}, f_{5})$, where $g(\cdot)$ denotes a multi-level feature aggregation algorithm. Previous works focus on how to develop a more effective integration strategy.

In Fig.~\ref{Structure}(b), we show the architecture of the proposed cascaded partial decoder framework. Since that the features of shallower layers contribute less to performance, we construct a partial decoder that only integrates features of deeper layers. In order to utilize generated saliency map to refine features, we design a bifurcated backbone network. We set the Conv$3\_3$ layer as an optimization layer, and use the last two convolutional blocks to construct two branches (an attention one and a detection one). In the attention branch, we design a partial decoder to integrate three-level features which are denoted as $\{f_{i}^{a}=f_{i}, i = 3,4,5\}$. Hence the partial decoder is represented by $D_{a} = g_{a}(f_{3}^{a}, f_{4}^{a}, f_{5}^{a})$ and it generates an initial saliency map $S_{i}$. After processing of the proposed holistic attention module, we obtain an enhanced attention map $S_{h}$ which is utilized to refine the feature $f_{3}$. Because we can obtain relatively precise saliency map via integrating features of three top layers, the attention map $S_{h}$ effectively eliminates distractors in feature $f_{3}$. Then we obtain the refined feature $f_{3}^{d}$ for detection branch via element-wise multiplying the feature and the attention map: $f_{3}^{d} = f_{3}\odot S_{h}$. Hence the following two-level features of the detection branch are denoted as $\{f_{4}^{d}, f_{5}^{d}\}$. Through constructing another partial decoder $D_{d} = g_{d}(f_{3}^{d}, f_{4}^{d}, f_{5}^{d})$ for the detection branch, the proposed model outputs the final saliency map $S_{d}$. For convenience, we set $g_{a} = g_{d}$ in this paper. The details of the proposed holistic attention module and the partial decoder are described in Section~\ref{3.2} and Section~\ref{3.3} respectively.

We jointly train the two branches with ground truth. The parameters of the two branches are not shared. Given $\{S_{i}, S_{d}\}$ and the corresponding label $l$, the total loss $L_{total}$ is formulated as:
\begin{equation}
  L_{total} = L_{ce}(S_{i}, l|\Theta_{i}) + L_{ce}(S_{d}, l|\Theta_{d}).
\end{equation}
$L_{ce}$ is the sigmoid cross entropy loss:
\begin{equation}
  L_{ce}(\Theta) = -\sum_{j=1}^{N}\sum_{c\in\{0,1\}}\delta(l^{j} = c)\log p(S^{j} = c|\Theta),
\end{equation}
where $N$ is the pixel number, $\delta$ is the indicator function, $j$ denotes pixel coordinate and $\Theta = \{\Theta_{i}, \Theta_{d}\}$ are parameter sets corresponding to the saliency maps $S = \{S_{i}, S_{d}\}$. It is obvious that $\Theta_{i}$ is a proper subset of $\Theta_{d}$, which indicates that the two branches work in an alternating way. On the one hand, the attention branch provides precise attention map for the detection branch, which leads to that the detection branch segments more accurate salient objects. On the other hand, the detection branch could be considered as an auxiliary loss of the attention branch, which also helps the attention branch to focus on salient objects. Joint training the two branches makes our model uniformly highlights salient objects while suppressing distractors.

In addition, we can leverage the proposed framework to improve existing deep aggregation models when we integrate the features of each branch by using the aggregation algorithms of these works. Even though we raise the computation cost of the backbone network and add one more decoder when compared to the traditional encoder-decoder architecture, the total computation complexity is still significantly reduced because of discarding low-level features in decoders. Moreover, the cascaded optimization mechanism of the proposed framework promotes the performance, and the experiments show that the two branches both outperform the original models.

\subsection{Holistic Attention Module}\label{3.2}
\begin{figure}[t]
\centering
\subfigure[Image]{
\begin{minipage}[b]{0.22\linewidth}
\includegraphics[width=1\linewidth]{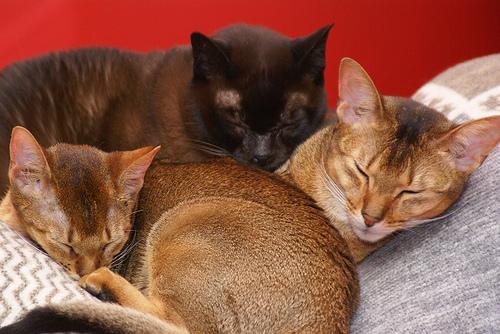}\vspace{2pt}
\includegraphics[width=1\linewidth]{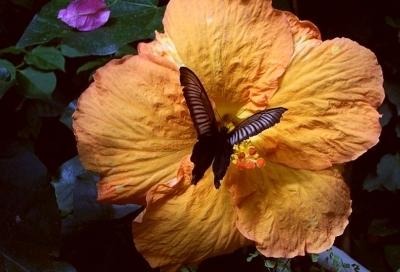}\vspace{2pt}
\includegraphics[width=1\linewidth]{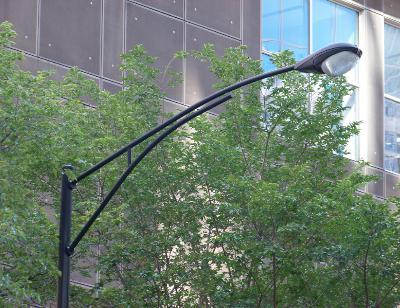}\vspace{2pt}
\end{minipage}}
\hspace{-1.2ex}
\subfigure[GT]{
\begin{minipage}[b]{0.22\linewidth}
\includegraphics[width=1\linewidth]{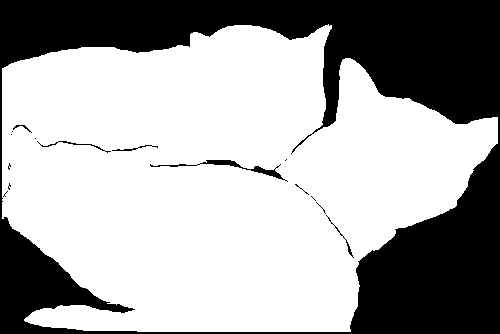}\vspace{2pt}
\includegraphics[width=1\linewidth]{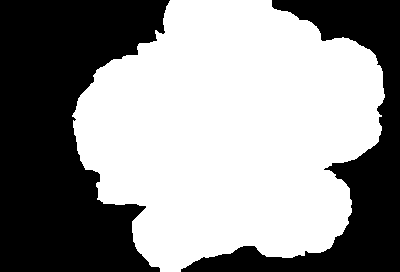}\vspace{2pt}
\includegraphics[width=1\linewidth]{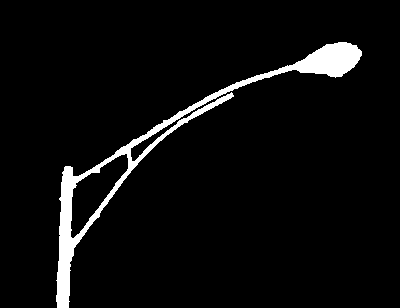}\vspace{2pt}
\end{minipage}}
\hspace{-1.2ex}
\subfigure[Initial Attention]{
\begin{minipage}[b]{0.22\linewidth}
\includegraphics[width=1\linewidth]{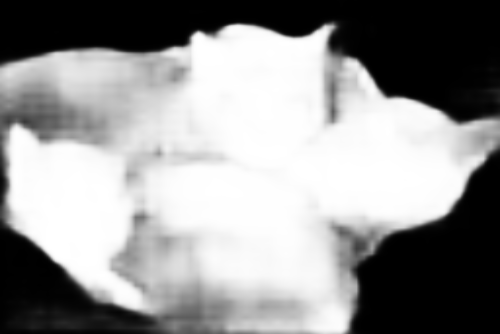}\vspace{2pt}
\includegraphics[width=1\linewidth]{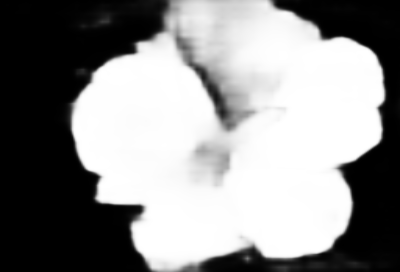}\vspace{2pt}
\includegraphics[width=1\linewidth]{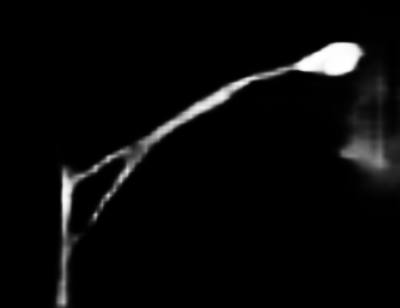}\vspace{2pt}
\end{minipage}}
\hspace{-1.2ex}
\subfigure[Holistic Attention]{
\begin{minipage}[b]{0.22\linewidth}
\includegraphics[width=1\linewidth]{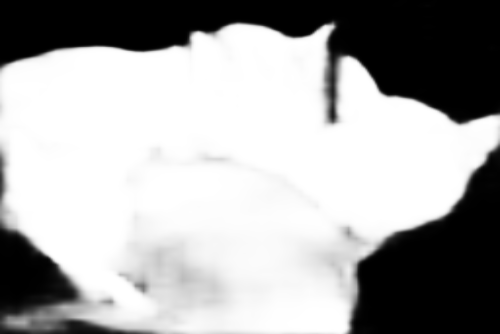}\vspace{2pt}
\includegraphics[width=1\linewidth]{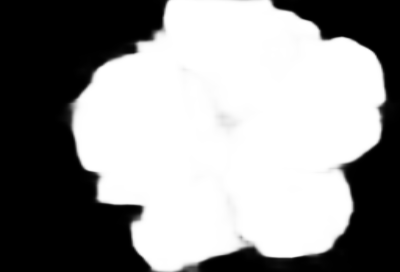}\vspace{2pt}
\includegraphics[width=1\linewidth]{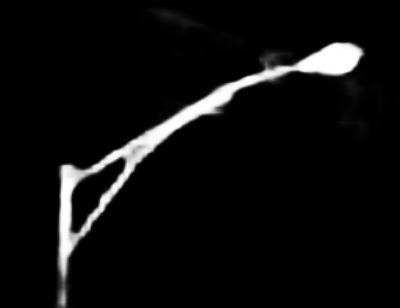}\vspace{2pt}
\end{minipage}}
\caption{\emph{GT is the ground truth. As we can see, the proposed holistic attention module is helpful for segmenting the whole salient objects and refining more precise boundaries.}}
\label{SoftAttentoinExamples}
\end{figure}

Given the feature map from the optimization layer and the initial saliency map from attention branch, we can use a initial attention strategy which means directly multiplying the feature map with the initial saliency map. When we obtain an accurate saliency map from the attention branch, this strategy will efficiently suppress distractors of the feature. On the contrary, if distractors are classified as salient regions, this strategy results in abnormal segmentation results. As a result, we need to improve the effectiveness of the initial saliency map. More specially, the edge information of salient objects may be filtered out by the initial saliency map because it is difficult to be precisely predicted. In addition, some objects in complex scenes are hard to be completely segmented. Therefore we propose a holistic attention module which aims to enlarge the coverage area of the initial saliency map, and it is defined as follows:
\begin{equation}
  S_{h} = MAX(f_{min\_max}(Conv_{g}(S_{i}, k)), S_{i})
\end{equation}
where $Conv_{g}$ is a convolution operation with a Gaussian kernel k and zero bias, $f_{min\_max}(\cdot)$ is a normalization function to make the blurred map ranges in $[0, 1]$, and $MAX(\cdot)$ is a maximum function which tends to increase the weight coefficient of salient regions of $S_{i}$ because that the convolution operation will blur $S_{i}$. Compared to the initial attention, the proposed holistic attention mechanism hardly increases computation cost, and it further highlights the whole salient objects as shown in Fig.~\ref{SoftAttentoinExamples}. Moreover, the size and standard deviation of Gaussian kernel k are initialized with $32$ and $4$. Then it is jointly trained with the proposed model.

\subsection{The Proposed Decoder}\label{3.3}
Since that the proposed framework consists of two decoders, we need to construct a fast integration strategy to ensure low complexity. Meanwhile, we need to generate saliency map as accurate as possible. Firstly, in order to capture global contrast information, we design an effective context module which is inspired by the receptive field block (RFB)~\cite{2018RFB}. Compared to the original RFB, we add one more branch to enlarge the receptive field further. Our context module consists of four branches $\{b_{m}, m=1,...,4\}$. For acceleration, in each branch, we use a $1\times 1$ convolutional layer to reduce channel number to $32$. For $\{b_{m}, m>1\}$, we add two layers: a $(2m-1)\times (2m-1)$ convolutional layer and a $3\times 3$ convolutional layer with $(2m-1)$ dilation~\cite{2018DeepLab}. We concatenate the outputs of these branches and reduce the channel to $32$ by an additional $1\times 1$ convolutional layer. Then a short connection is added as the original RFB. In general, given features $\{f_{i}^{c}, i\in[l,...,L], c\in[a, d]\}$ from the bifurcated backbone network, we obtain discriminative features $\{f_{i}^{c_{1}}\}$ from the context module. Then we use multiplication operation to reduce the gap between multi-level features. Especially, for the top-most feature ($i = L$), we set $f_{L}^{c_{2}} = f_{L}^{c_{1}}$. For feature $\{f_{i}^{c_{1}}, i<L\}$, we update it to $f_{i}^{c_{2}}$ via element-wise multiplying itself with all features of deeper layers. This operation is defined as follows:
\begin{equation}
  f_{i}^{c_{2}} = f_{i}^{c_{1}}\odot \Pi_{k=i+1}^{L}Conv(Up(f_{k}^{c_{1}})), i\in[l,...,L-1],
\end{equation}
where $Up(\cdot)$ is upsampling feature by a factor $2^{k-j}$, and $Conv$ is a $3\times 3$ convolutional layer. At last, we utilize an upsampling-concatenating strategy to integrate multi-level features. When we construct a partial decoder and set the Conv$3\_3$ layer as the optimization layer ($l=3, L=5$), we obtain a feature map with $[\frac{H}{4}, \frac{W}{4}]$ size and $96$ channel number. With $3\times 3$ layer and $1\times 1$ convolutional layers, we obtain the final feature map and resize it to $[H, W]$.

\section{Experiment}
\subsection{Salient Object Detection}
\subsubsection{Experimental Setup}
\textbf{Evaluation Datasets. }We evaluate the proposed model on five benchmark datasets: ECSSD~\cite{Dataset-ECSSD}, HKU-IS~\cite{2015MDF}, PASCAL-S~\cite{Dataset-PASCAL-S}, DUTS~\cite{Dataset-DUTS}, DUT-OMRON~\cite{Dataset-DUT-OMRON}.

\begin{table*}[http]
  \centering
  \resizebox{\textwidth}{!}{
    \begin{tabular}{c|c|c|ccc|ccc|ccc|ccc|ccc}
    \hline
    \multicolumn{1}{c|}{\multirow{2}[0]{*}{Method}} & \multicolumn{1}{c|}{\multirow{2}[0]{*}{Backbone}} & \multicolumn{1}{c|}{\multirow{2}[0]{*}{FPS}} & \multicolumn{3}{c|}{ECSSD~\cite{Dataset-ECSSD}} & \multicolumn{3}{c|}{HKU-IS~\cite{2015MDF}} & \multicolumn{3}{c|}{DUT-OMRON~\cite{Dataset-DUT-OMRON}} & \multicolumn{3}{c|}{DUTS~\cite{Dataset-DUTS}} & \multicolumn{3}{c}{PASCAL-S~\cite{Dataset-PASCAL-S}} \\
    \cline{4-18}
          &       &       & maxF & avgF & MAE & maxF & avgF & MAE & maxF & avgF & MAE & maxF & avgF & MAE & maxF & avgF & MAE \\
    \hline
    Amulet~\cite{2017Amulet} & VGG16 & 21 & 0.922 & 0.881 & 0.057 & 0.909 & 0.863 & 0.047 & 0.791 & 0.699 & 0.072 & 0.832 & 0.738 & 0.062 & 0.839 & 0.780 & 0.095 \\
    NLDF~\cite{2017NLDF}  & VGG16 & 20 & 0.915 & 0.886 & 0.051 & 0.908 & 0.871 & 0.041 & 0.759 & 0.694 & 0.071 & 0.830 & 0.759 & 0.055 & 0.840 & 0.792 & 0.083 \\
    DSS~\cite{2018DSS}   & VGG16 & 23 & 0.928 & 0.889 & 0.051 & 0.915 & 0.867 & 0.043 & 0.781 & 0.692 & 0.065 & 0.858 & 0.757 & 0.050 & 0.859 & 0.796 & 0.081 \\
    BMPM~\cite{2018BMPM}  & VGG16 & 28 & 0.928 & 0.894 & 0.044 & 0.920 & 0.875 & 0.039 & 0.775 & 0.693 & 0.063 & 0.850 & 0.768 & 0.049 & 0.862 & 0.770 & \textbf{{\color{red}0.074}} \\
    PAGR~\cite{2018PAGR}  & VGG19 & - & 0.927 & 0.894 & 0.061 & 0.918 & 0.886 & 0.048 & 0.771 & 0.711 & 0.072 & 0.855 & 0.788 & 0.055 & 0.851 & 0.803 & 0.092 \\
    PiCANet~\cite{2018PiCANet} & VGG16 & 7 & 0.931 & 0.885 & 0.046 & 0.921 & 0.870 & 0.042 & \textbf{{\color{red}0.794}} & 0.710 & 0.068 & 0.851 & 0.749 & 0.054 & 0.862 & 0.796 & 0.076 \\
    \textbf{\emph{CPD-A}} (ours) & VGG16 & \textbf{{\color{red}105}} & 0.928 & 0.906 & 0.045 & 0.918 & 0.884 & 0.037 & 0.781 & 0.721 & 0.061 & 0.854 & 0.787 & 0.047 & 0.859 & 0.814 & 0.077 \\
    \textbf{\emph{CPD}} (ours)   & VGG16 & 66 & \textbf{{\color{red}0.936}} & \textbf{{\color{red}0.915}} & \textbf{{\color{red}0.040}} & \textbf{{\color{red}0.924}} & \textbf{{\color{red}0.896}} & \textbf{{\color{red}0.033}} & \textbf{{\color{red}0.794}} & \textbf{{\color{red}0.745}} & \textbf{{\color{red}0.057}} & \textbf{{\color{red}0.864}} & \textbf{{\color{red}0.813}} & \textbf{{\color{red}0.043}} & \textbf{{\color{red}0.866}} & \textbf{{\color{red}0.825}} & \textbf{{\color{red}0.074}} \\
    \hline
    \hline
    SRM~\cite{2017SRM}   & ResNet50 & 37 & 0.917 & 0.892 & 0.054 & 0.903 & 0.871 & 0.047 & 0.769 & 0.707 & 0.069 & 0.827 & 0.757 & 0.059 & 0.847 & 0.796 & 0.085 \\
    DGRL~\cite{2018DGRL}  & ResNet50 & 6 & 0.925 & 0.903 & 0.043 & 0.914 & 0.882 & 0.037 & 0.779 & 0.709 & 0.063 & 0.834 & 0.764 & 0.051 & 0.853 & 0.807 & 0.074 \\
    PiCANet-R~\cite{2018PiCANet} & ResNet50 & 5 & 0.935 & 0.886 & 0.046 & 0.919 & 0.870 & 0.043 & \textbf{{\color{red}0.803}} & 0.717 & 0.065 & 0.860 & 0.759 & 0.051 & 0.863 & 0.798 & 0.075 \\
    \textbf{\emph{CPD-RA}} (ours) & ResNet50 & \textbf{{\color{red}104}} & 0.934 & 0.907 & 0.043 & 0.918 & 0.882 & 0.038 & 0.783 & 0.725 & 0.059 & 0.852 & 0.776 & 0.048 & 0.855 & 0.807 & 0.077 \\
    \textbf{\emph{CPD-R}} (ours) & ResNet50 & 62 & \textbf{{\color{red}0.939}} & \textbf{{\color{red}0.917}} & \textbf{{\color{red}0.037}} & \textbf{{\color{red}0.925}} & \textbf{{\color{red}0.891}} & \textbf{{\color{red}0.034}} & 0.797 & \textbf{{\color{red}0.747}} & \textbf{{\color{red}0.056}} & \textbf{{\color{red}0.865}} & \textbf{{\color{red}0.805}} & \textbf{{\color{red}0.043}} & \textbf{{\color{red}0.864}} & \textbf{{\color{red}0.824}} & \textbf{{\color{red}0.072}} \\
    \hline
    \end{tabular}}
  \caption{\emph{Comparison of different methods on five benchmark datasets and four metrics including FPS, MAE (lower is better), max F-measure (higher is better) and average F-measure. The comparison is under two settings (with VGG~\cite{2014VGG} and ResNet50~\cite{2016ResNet} backbone netowrk). The best result of each setting is shown in \textbf{{\color{red}Red}}. ``-R" means using ResNet$50$ as the backbone. ``-A" means the results of the attention branch. All method are the trained on training set of DUTS~\cite{Dataset-DUTS}. There is not available code of PAGR~\cite{2018PAGR} and the author only provides the saliency maps.}}
  \label{All_Results}%
\end{table*}%

\begin{figure*}[http]
\centering
\subfigure[\tiny Image]{
\begin{minipage}[b]{0.062\linewidth}
\includegraphics[width=1\linewidth]{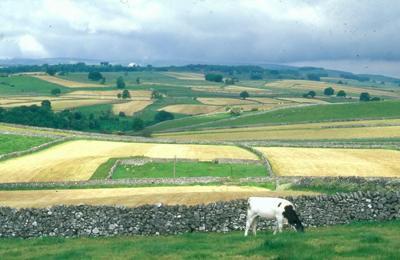}\vspace{2pt}
\includegraphics[width=1\linewidth]{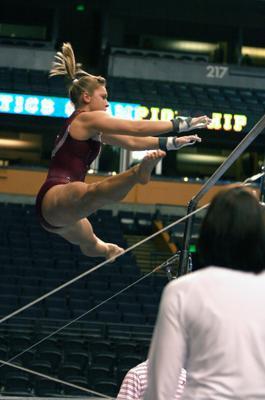}\vspace{2pt}
\includegraphics[width=1\linewidth]{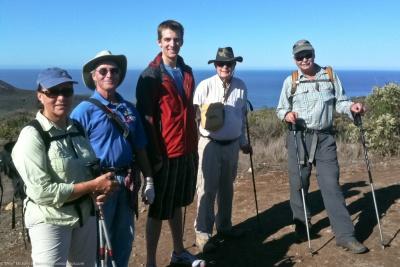}\vspace{2pt}
\includegraphics[width=1\linewidth]{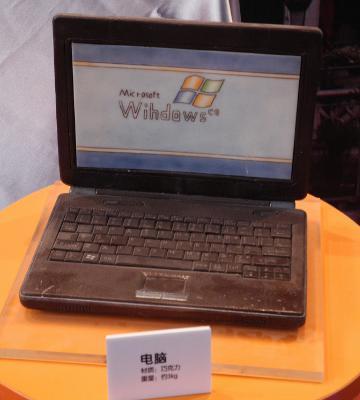}\vspace{2pt}
\end{minipage}}
\hspace{-1.2ex}
\subfigure[\tiny GT]{
\begin{minipage}[b]{0.062\linewidth}
\includegraphics[width=1\linewidth]{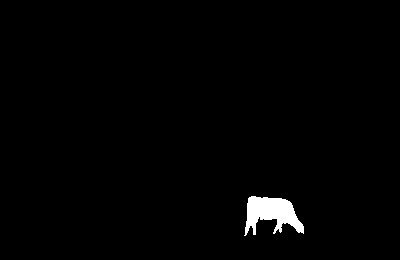}\vspace{2pt}
\includegraphics[width=1\linewidth]{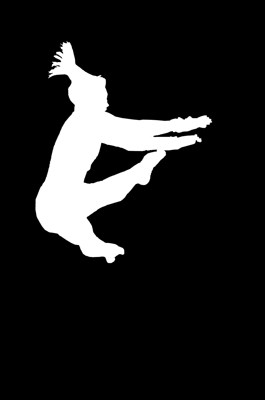}\vspace{2pt}
\includegraphics[width=1\linewidth]{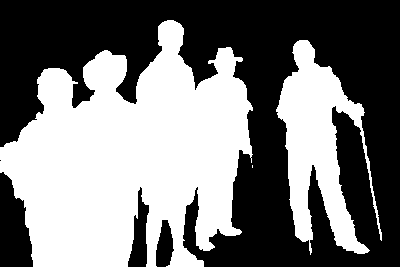}\vspace{2pt}
\includegraphics[width=1\linewidth]{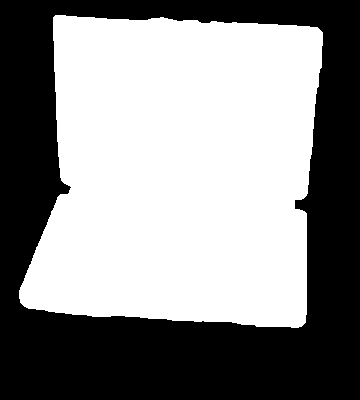}\vspace{2pt}
\end{minipage}}
\hspace{-1.2ex}
\subfigure[\tiny \textbf{\emph{CPD}}]{
\begin{minipage}[b]{0.062\linewidth}
\includegraphics[width=1\linewidth]{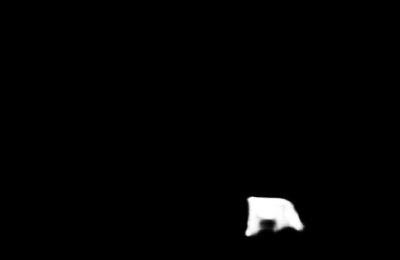}\vspace{2pt}
\includegraphics[width=1\linewidth]{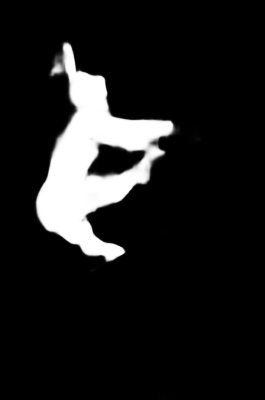}\vspace{2pt}
\includegraphics[width=1\linewidth]{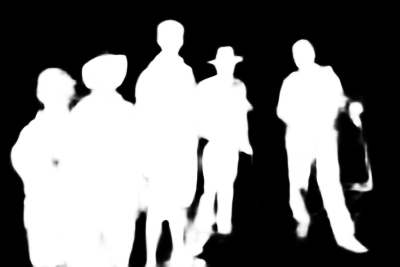}\vspace{2pt}
\includegraphics[width=1\linewidth]{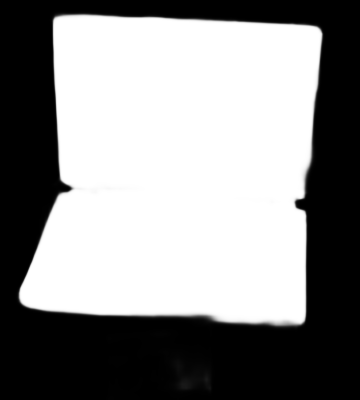}\vspace{2pt}
\end{minipage}}
\hspace{-1.2ex}
\subfigure[\tiny \textbf{\emph{CPD-R}}]{
\begin{minipage}[b]{0.062\linewidth}
\includegraphics[width=1\linewidth]{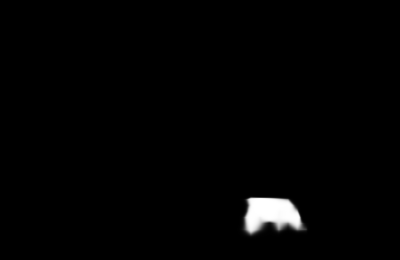}\vspace{2pt}
\includegraphics[width=1\linewidth]{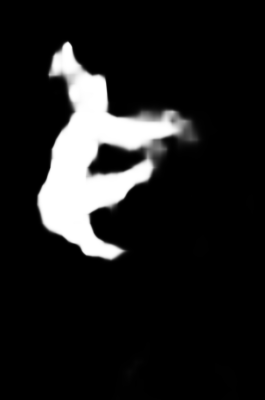}\vspace{2pt}
\includegraphics[width=1\linewidth]{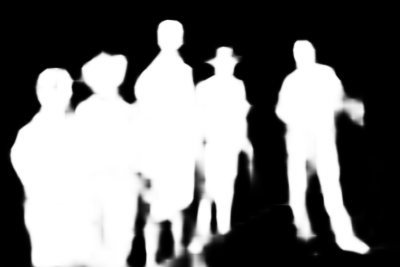}\vspace{2pt}
\includegraphics[width=1\linewidth]{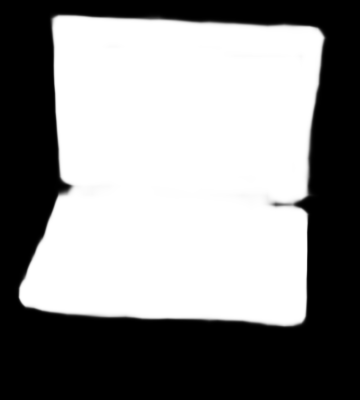}\vspace{2pt}
\end{minipage}}
\hspace{-1.2ex}
\subfigure[\tiny \textbf{\emph{CPD-A}}]{
\begin{minipage}[b]{0.062\linewidth}
\includegraphics[width=1\linewidth]{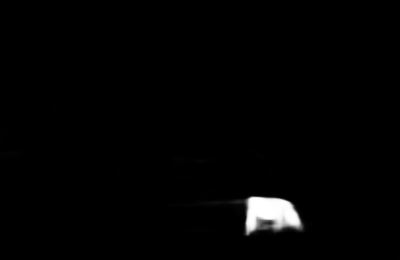}\vspace{2pt}
\includegraphics[width=1\linewidth]{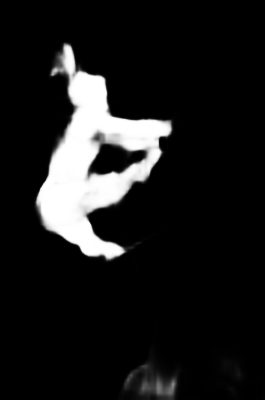}\vspace{2pt}
\includegraphics[width=1\linewidth]{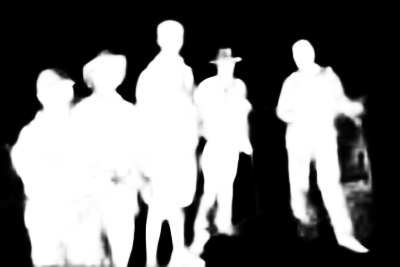}\vspace{2pt}
\includegraphics[width=1\linewidth]{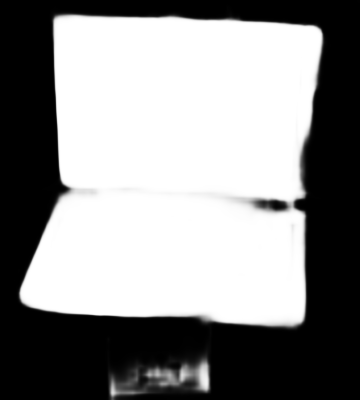}\vspace{2pt}
\end{minipage}}
\hspace{-1.2ex}
\subfigure[\tiny \textbf{\emph{CPD-RA}}]{
\begin{minipage}[b]{0.062\linewidth}
\includegraphics[width=1\linewidth]{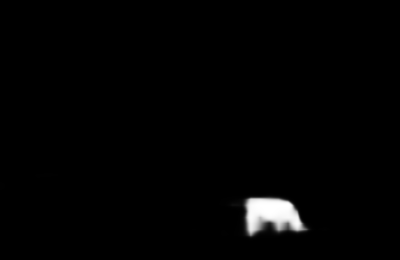}\vspace{2pt}
\includegraphics[width=1\linewidth]{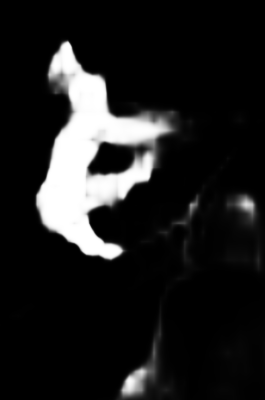}\vspace{2pt}
\includegraphics[width=1\linewidth]{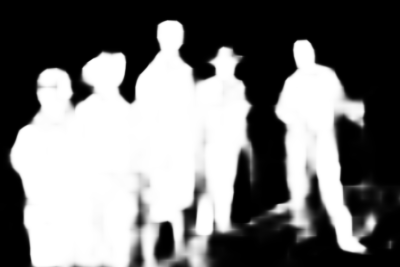}\vspace{2pt}
\includegraphics[width=1\linewidth]{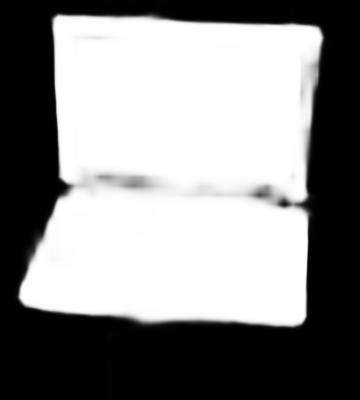}\vspace{2pt}
\end{minipage}}
\hspace{-1.2ex}
\subfigure[\tiny PiCANet-R]{
\begin{minipage}[b]{0.062\linewidth}
\includegraphics[width=1\linewidth]{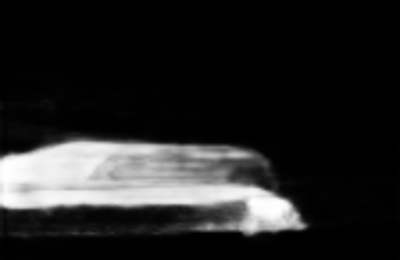}\vspace{2pt}
\includegraphics[width=1\linewidth]{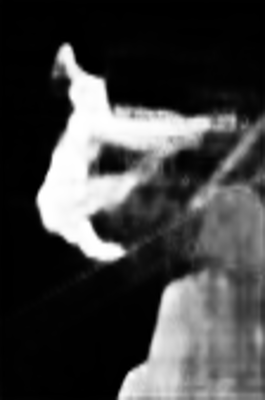}\vspace{2pt}
\includegraphics[width=1\linewidth]{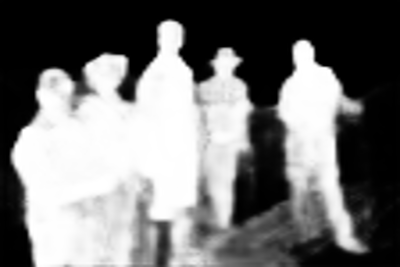}\vspace{2pt}
\includegraphics[width=1\linewidth]{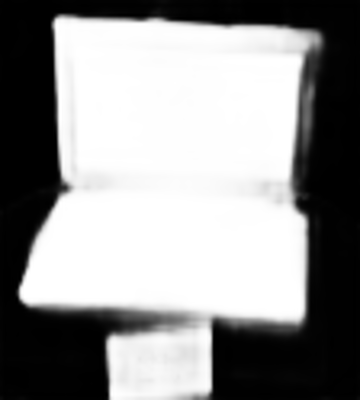}\vspace{2pt}
\end{minipage}}
\hspace{-1.2ex}
\subfigure[\tiny PiCANet]{
\begin{minipage}[b]{0.062\linewidth}
\includegraphics[width=1\linewidth]{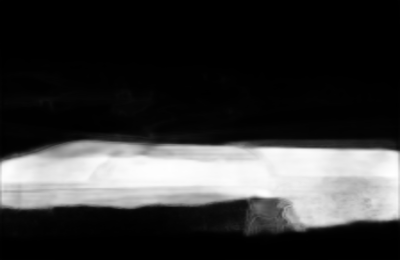}\vspace{2pt}
\includegraphics[width=1\linewidth]{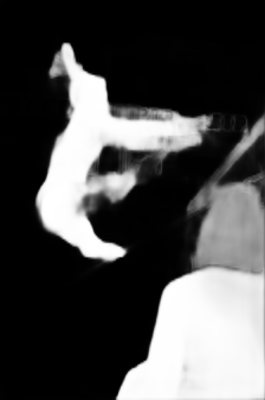}\vspace{2pt}
\includegraphics[width=1\linewidth]{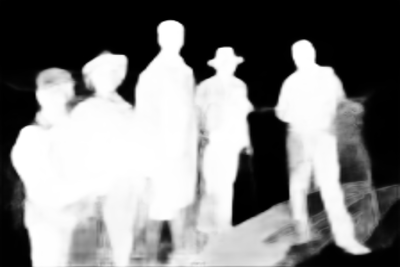}\vspace{2pt}
\includegraphics[width=1\linewidth]{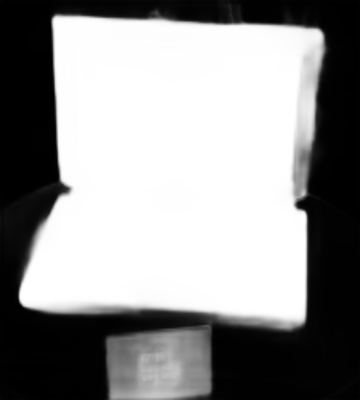}\vspace{2pt}
\end{minipage}}
\hspace{-1.2ex}
\subfigure[\tiny PAGR]{
\begin{minipage}[b]{0.062\linewidth}
\includegraphics[width=1\linewidth]{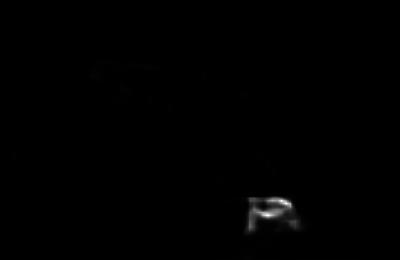}\vspace{2pt}
\includegraphics[width=1\linewidth]{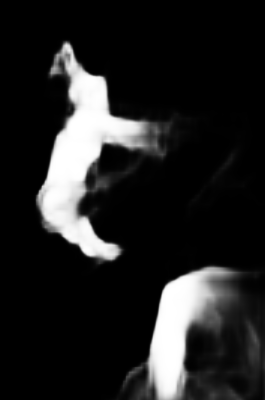}\vspace{2pt}
\includegraphics[width=1\linewidth]{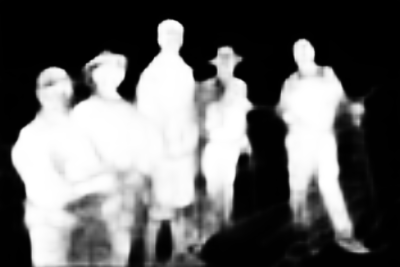}\vspace{2pt}
\includegraphics[width=1\linewidth]{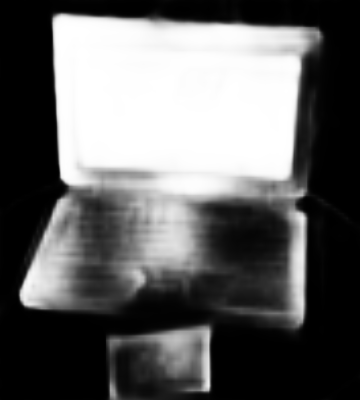}\vspace{2pt}
\end{minipage}}
\hspace{-1.2ex}
\subfigure[\tiny DGRL]{
\begin{minipage}[b]{0.062\linewidth}
\includegraphics[width=1\linewidth]{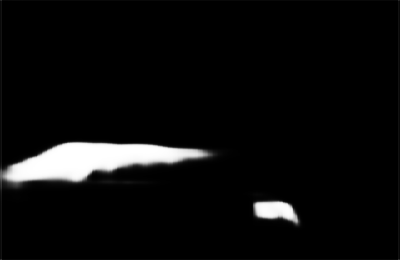}\vspace{2pt}
\includegraphics[width=1\linewidth]{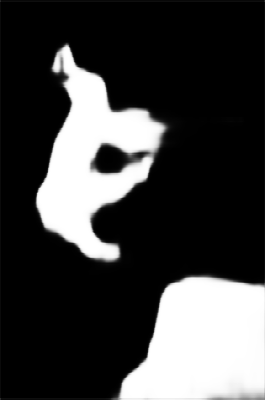}\vspace{2pt}
\includegraphics[width=1\linewidth]{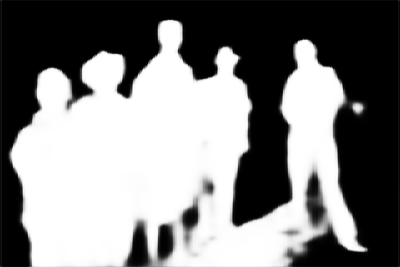}\vspace{2pt}
\includegraphics[width=1\linewidth]{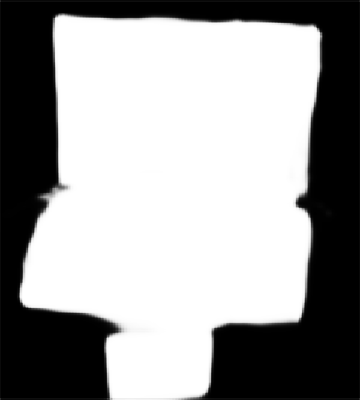}\vspace{2pt}
\end{minipage}}
\hspace{-1.2ex}
\subfigure[\tiny BMPM]{
\begin{minipage}[b]{0.062\linewidth}
\includegraphics[width=1\linewidth]{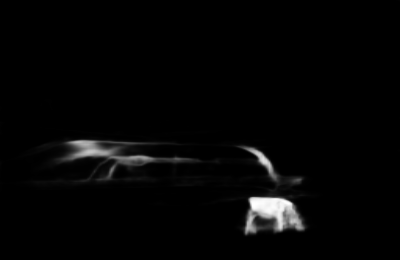}\vspace{2pt}
\includegraphics[width=1\linewidth]{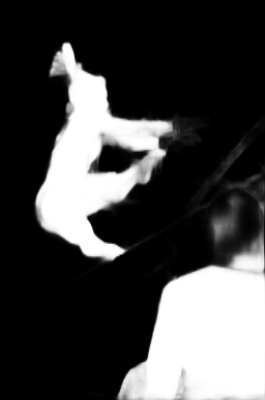}\vspace{2pt}
\includegraphics[width=1\linewidth]{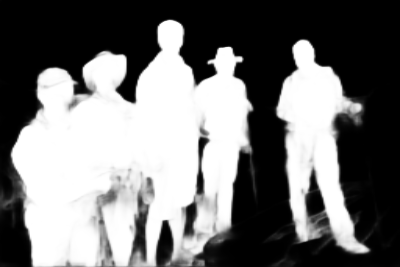}\vspace{2pt}
\includegraphics[width=1\linewidth]{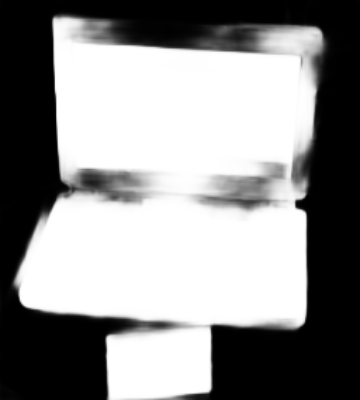}\vspace{2pt}
\end{minipage}}
\hspace{-1.2ex}
\subfigure[\tiny SRM]{
\begin{minipage}[b]{0.062\linewidth}
\includegraphics[width=1\linewidth]{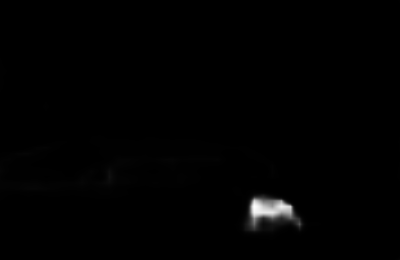}\vspace{2pt}
\includegraphics[width=1\linewidth]{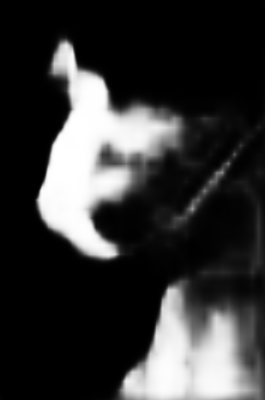}\vspace{2pt}
\includegraphics[width=1\linewidth]{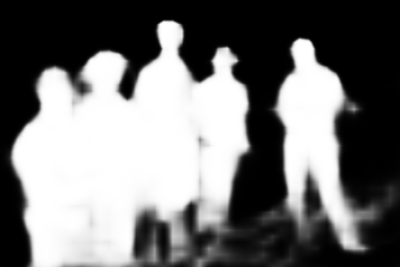}\vspace{2pt}
\includegraphics[width=1\linewidth]{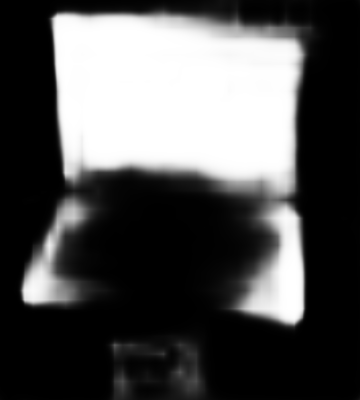}\vspace{2pt}
\end{minipage}}
\hspace{-1.2ex}
\subfigure[\tiny DSS]{
\begin{minipage}[b]{0.062\linewidth}
\includegraphics[width=1\linewidth]{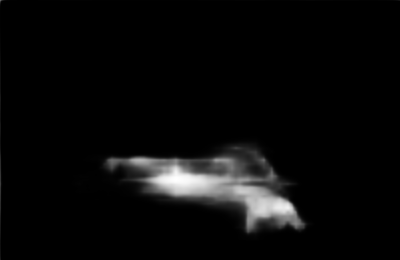}\vspace{2pt}
\includegraphics[width=1\linewidth]{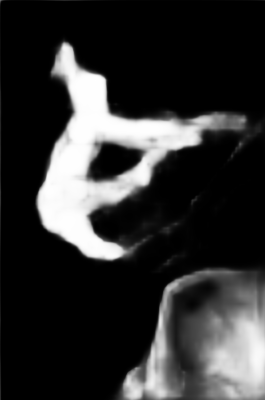}\vspace{2pt}
\includegraphics[width=1\linewidth]{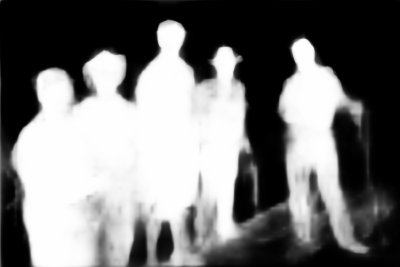}\vspace{2pt}
\includegraphics[width=1\linewidth]{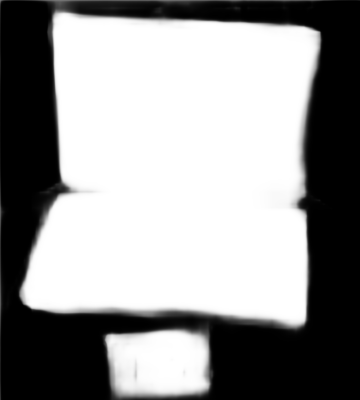}\vspace{2pt}
\end{minipage}}
\hspace{-1.2ex}
\subfigure[\tiny Amulet]{
\begin{minipage}[b]{0.062\linewidth}
\includegraphics[width=1\linewidth]{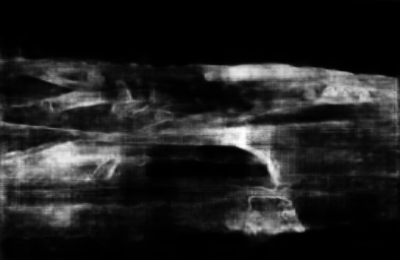}\vspace{2pt}
\includegraphics[width=1\linewidth]{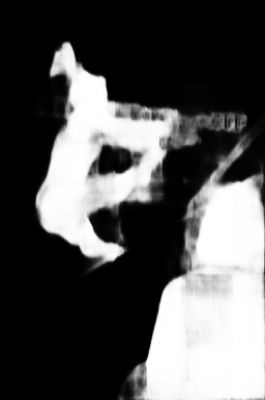}\vspace{2pt}
\includegraphics[width=1\linewidth]{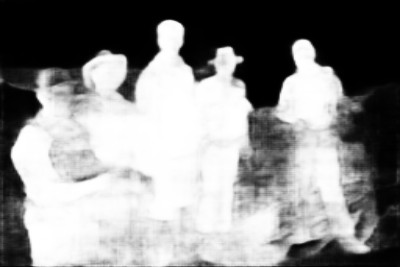}\vspace{2pt}
\includegraphics[width=1\linewidth]{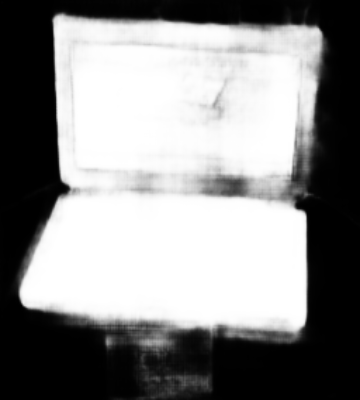}\vspace{2pt}
\end{minipage}}
\hspace{-1.2ex}
\subfigure[\tiny NLDF]{
\begin{minipage}[b]{0.062\linewidth}
\includegraphics[width=1\linewidth]{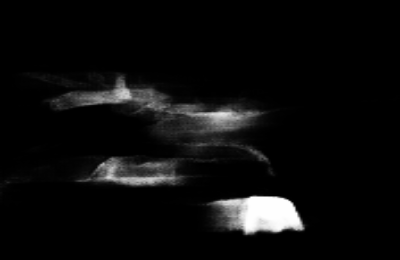}\vspace{2pt}
\includegraphics[width=1\linewidth]{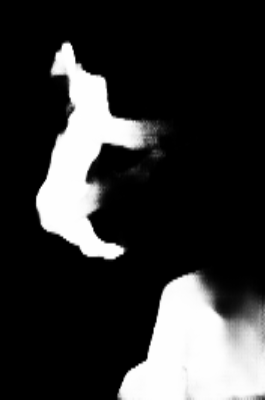}\vspace{2pt}
\includegraphics[width=1\linewidth]{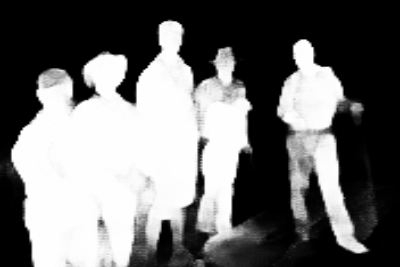}\vspace{2pt}
\includegraphics[width=1\linewidth]{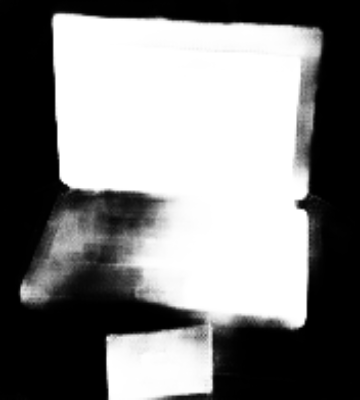}\vspace{2pt}
\end{minipage}}
\caption{\emph{Visual comparisons of the proposed model and existing state-of-the-art algorithms in some challenging cases: small object, complex scene, multiple objects and large object.}}
\label{All_Figures}
\end{figure*}

\begin{table*}[ht]
  \centering
  \resizebox{\textwidth}{!}{
    \begin{tabular}{c|c|ccc|ccc|ccc|ccc|ccc}
      \hline
         \multicolumn{1}{c|}{\multirow{2}[0]{*}{Method}} & \multicolumn{1}{c|}{\multirow{2}[0]{*}{FPS}} & \multicolumn{3}{c|}{ECSSD~\cite{Dataset-ECSSD}} & \multicolumn{3}{c|}{HKU-IS~\cite{2015MDF}} & \multicolumn{3}{c|}{DUT-OMRON~\cite{Dataset-DUT-OMRON}} & \multicolumn{3}{c|}{DUTS~\cite{Dataset-DUTS}} & \multicolumn{3}{c}{PASCAL-S~\cite{Dataset-PASCAL-S}} \\
      \cline{3-17}
          &       & maxF & avgF & MAE & maxF  & avgF & MAE & maxF & avgF & MAE & maxF & avgF & MAE & maxF & avgF & MAE\\
      \hline
    BMPM~\cite{2018BMPM}  & 28 & 0.928 & 0.894 & 0.044 & 0.920 & 0.875 & 0.039 & 0.775 & 0.693 & 0.063 & 0.850 & 0.768 & 0.049 & 0.862 & 0.803 & 0.074 \\
    \textbf{\emph{BMPM-CPD-A}} & \textbf{{\color{red}82}} & 0.932 & 0.901 & 0.046 & 0.920 & 0.882 & 0.037 & 0.796 & 0.731 & 0.057 & 0.864 & 0.799 & 0.046 & 0.861 & 0.817 & 0.074 \\
    \textbf{\emph{BMPM-CPD}} & 47 & \textbf{{\color{red}0.935}} & \textbf{{\color{red}0.907}} & \textbf{{\color{red}0.043}} & \textbf{{\color{red}0.925}} & \textbf{{\color{red}0.888}} & \textbf{{\color{red}0.035}} & \textbf{{\color{red}0.804}} & \textbf{{\color{red}0.740}} & \textbf{{\color{red}0.056}} & \textbf{{\color{red}0.870}} & \textbf{{\color{red}0.808}} & \textbf{{\color{red}0.044}} & \textbf{{\color{red}0.868}} & \textbf{{\color{red}0.822}} & \textbf{{\color{red}0.072}} \\
    \hline
    \hline
    NLDF~\cite{2017NLDF}  & 21 & 0.915 & 0.886 & 0.051 & 0.908 & 0.871 & 0.041 & 0.759 & 0.694 & 0.071 & 0.830 & 0.759 & 0.055 & 0.840 & 0.792 & 0.083 \\
    \textbf{\emph{NLDF-CPD-A}} & \textbf{{\color{red}75}} & 0.918 & 0.889 & 0.049 & 0.914 & 0.873 & 0.039 & 0.775 & 0.710 & 0.061 & 0.837 & 0.773 & 0.050 & 0.841 & 0.793 & 0.083 \\
    \textbf{\emph{NLDF-CPD}} & 48 & \textbf{{\color{red}0.922}} & \textbf{{\color{red}0.896}} & \textbf{{\color{red}0.044}} & \textbf{{\color{red}0.916}} & \textbf{{\color{red}0.880}} & \textbf{{\color{red}0.036}} & \textbf{{\color{red}0.781}} & \textbf{{\color{red}0.721}} & \textbf{{\color{red}0.060}} & \textbf{{\color{red}0.842}} & \textbf{{\color{red}0.786}} & \textbf{{\color{red}0.048}} & \textbf{{\color{red}0.843}} & \textbf{{\color{red}0.800}} & \textbf{{\color{red}0.080}} \\
    \hline
    \hline
    Amulet~\cite{2017Amulet}  & 21 & 0.922 & 0.881 & 0.057 & 0.909 & 0.863 & 0.047 & 0.791 & 0.699 & 0.072 & 0.832 & 0.738 & 0.062 & 0.839 & 0.780 & 0.095 \\
    \textbf{\emph{Amulet-CPD-A}} & \textbf{{\color{red}61}} & 0.925 & 0.889 & 0.053 & 0.910 & 0.864 & 0.045 & 0.790 & 0.708 & 0.070 & 0.832 & 0.747 & 0.060 & 0.842 & 0.784 & 0.091 \\
    \textbf{\emph{Amulet-CPD}} & 45 & \textbf{{\color{red}0.934}} & \textbf{{\color{red}0.901}} & \textbf{{\color{red}0.047}} & \textbf{{\color{red}0.920}} & \textbf{{\color{red}0.878}} & \textbf{{\color{red}0.040}} & \textbf{{\color{red}0.805}} & \textbf{{\color{red}0.735}} & \textbf{{\color{red}0.063}} & \textbf{{\color{red}0.845}} & \textbf{{\color{red}0.771}} & \textbf{{\color{red}0.055}} & \textbf{{\color{red}0.851}} & \textbf{{\color{red}0.801}} & \textbf{{\color{red}0.085}} \\
    \hline
    \end{tabular}}
  \caption{\emph{Comparison of the original models and the improved models (-CPD-A and -CPD).}}
  \label{CPD_IN_OtherModels}
\end{table*}

\begin{figure*}[ht]
\centering
\subfigure[\tiny Image]{
\begin{minipage}[b]{0.087\linewidth}
\includegraphics[width=1\linewidth]{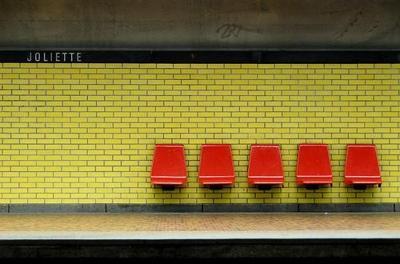}\vspace{2pt}
\includegraphics[width=1\linewidth]{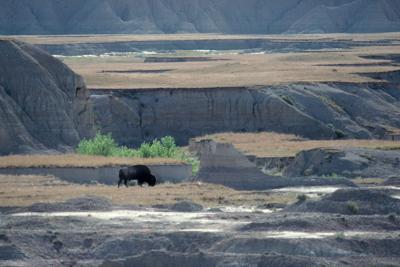}\vspace{2pt}
\includegraphics[width=1\linewidth]{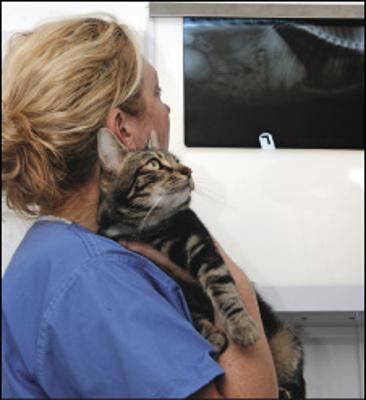}\vspace{2pt}
\includegraphics[width=1\linewidth]{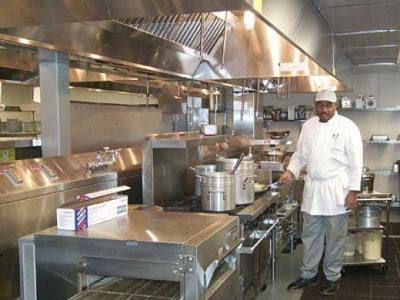}\vspace{2pt}
\end{minipage}}
\hspace{-1.2ex}
\subfigure[\tiny GT]{
\begin{minipage}[b]{0.087\linewidth}
\includegraphics[width=1\linewidth]{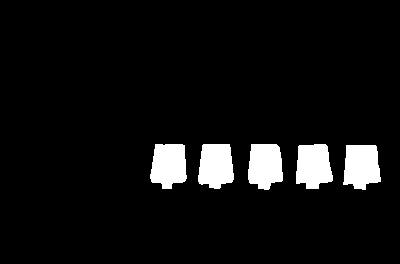}\vspace{2pt}
\includegraphics[width=1\linewidth]{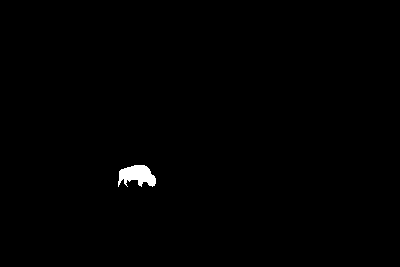}\vspace{2pt}
\includegraphics[width=1\linewidth]{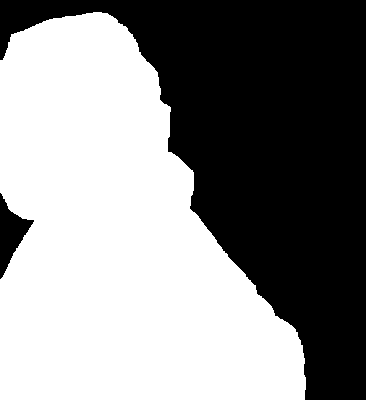}\vspace{2pt}
\includegraphics[width=1\linewidth]{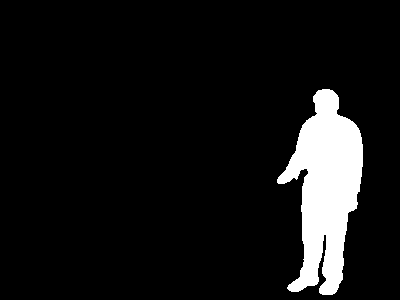}\vspace{2pt}
\end{minipage}}
\hspace{-1.2ex}
\subfigure[\tiny BMPM]{
\begin{minipage}[b]{0.087\linewidth}
\includegraphics[width=1\linewidth]{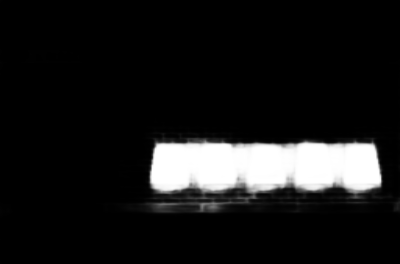}\vspace{2pt}
\includegraphics[width=1\linewidth]{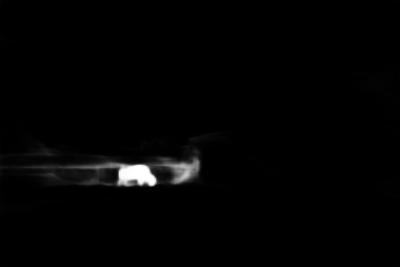}\vspace{2pt}
\includegraphics[width=1\linewidth]{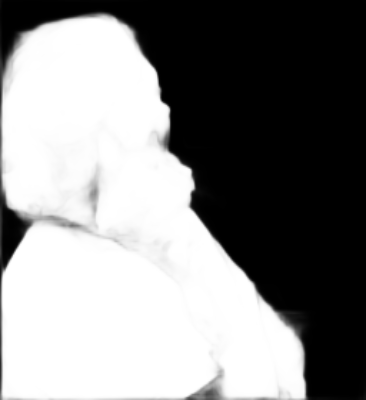}\vspace{2pt}
\includegraphics[width=1\linewidth]{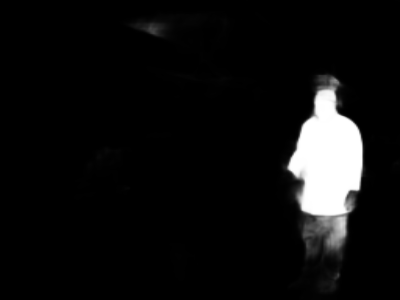}\vspace{2pt}
\end{minipage}}
\hspace{-1.2ex}
\subfigure[\tiny \textbf{\emph{BMPM-CPD-A}}]{
\begin{minipage}[b]{0.087\linewidth}
\includegraphics[width=1\linewidth]{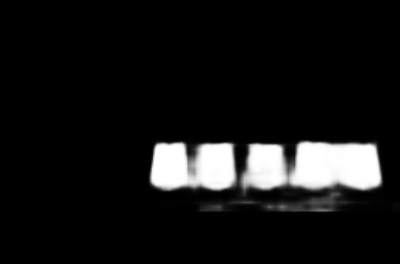}\vspace{2pt}
\includegraphics[width=1\linewidth]{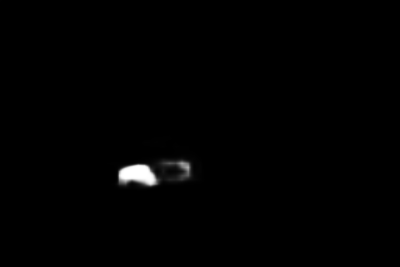}\vspace{2pt}
\includegraphics[width=1\linewidth]{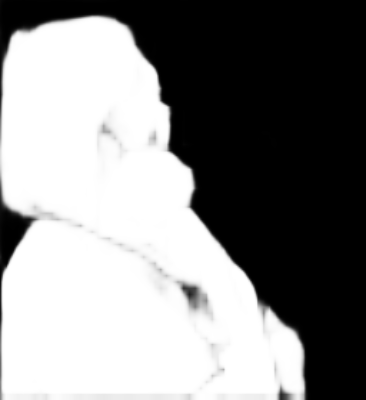}\vspace{2pt}
\includegraphics[width=1\linewidth]{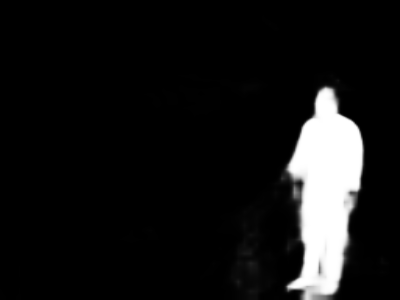}\vspace{2pt}
\end{minipage}}
\hspace{-1.2ex}
\subfigure[\tiny \textbf{\emph{BMPM-CPD}}]{
\begin{minipage}[b]{0.087\linewidth}
\includegraphics[width=1\linewidth]{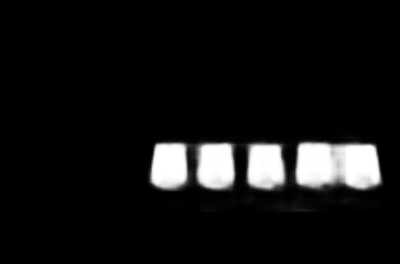}\vspace{2pt}
\includegraphics[width=1\linewidth]{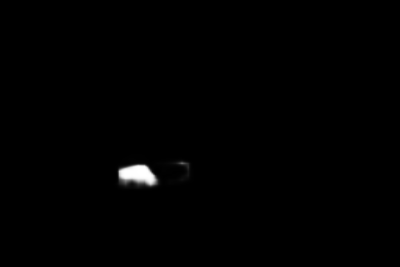}\vspace{2pt}
\includegraphics[width=1\linewidth]{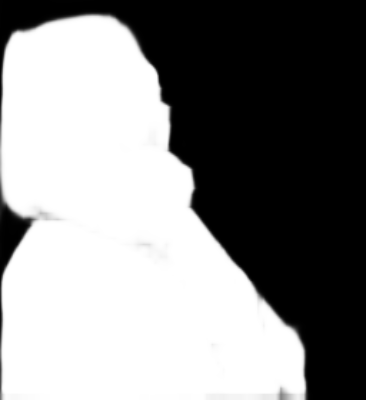}\vspace{2pt}
\includegraphics[width=1\linewidth]{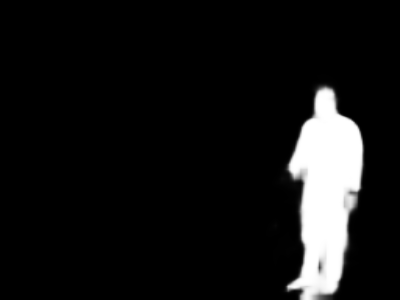}\vspace{2pt}
\end{minipage}}
\hspace{-1.2ex}
\subfigure[\tiny Amulet]{
\begin{minipage}[b]{0.087\linewidth}
\includegraphics[width=1\linewidth]{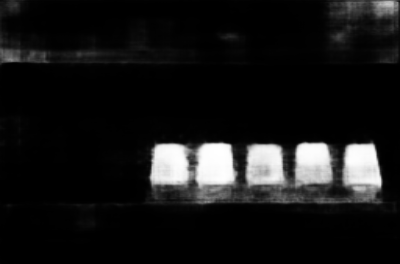}\vspace{2pt}
\includegraphics[width=1\linewidth]{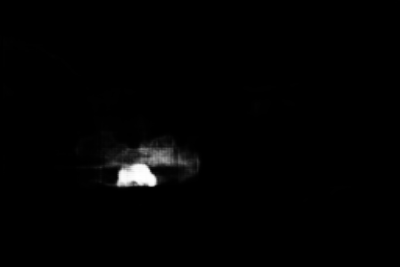}\vspace{2pt}
\includegraphics[width=1\linewidth]{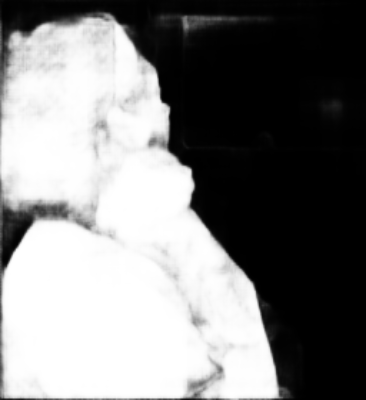}\vspace{2pt}
\includegraphics[width=1\linewidth]{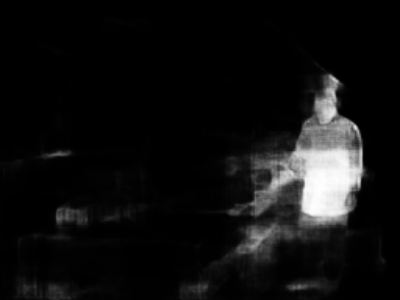}\vspace{2pt}
\end{minipage}}
\hspace{-1.2ex}
\subfigure[\tiny \textbf{\emph{Amulet-CPD-A}}]{
\begin{minipage}[b]{0.087\linewidth}
\includegraphics[width=1\linewidth]{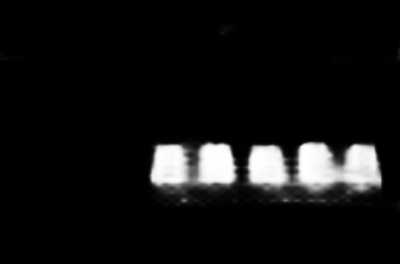}\vspace{2pt}
\includegraphics[width=1\linewidth]{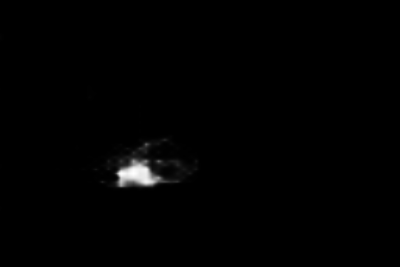}\vspace{2pt}
\includegraphics[width=1\linewidth]{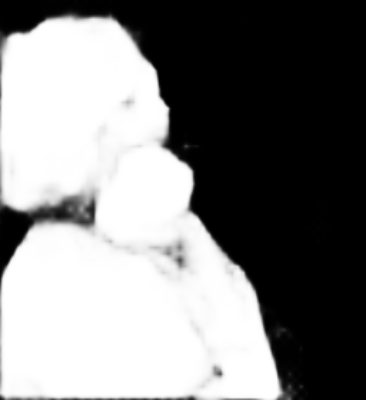}\vspace{2pt}
\includegraphics[width=1\linewidth]{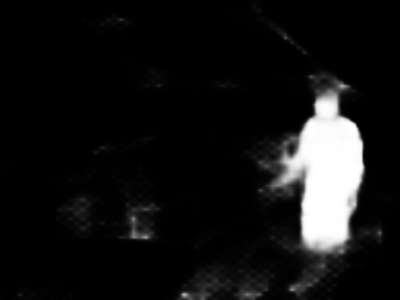}\vspace{2pt}
\end{minipage}}
\hspace{-1.2ex}
\subfigure[\tiny \textbf{\emph{Amulet-CPD}}]{
\begin{minipage}[b]{0.087\linewidth}
\includegraphics[width=1\linewidth]{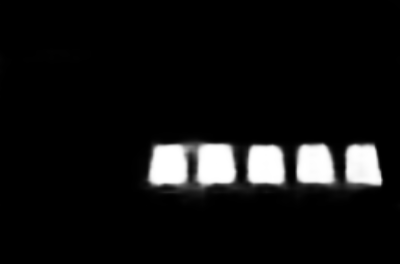}\vspace{2pt}
\includegraphics[width=1\linewidth]{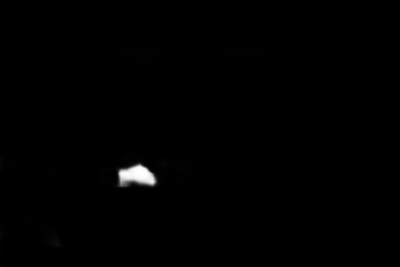}\vspace{2pt}
\includegraphics[width=1\linewidth]{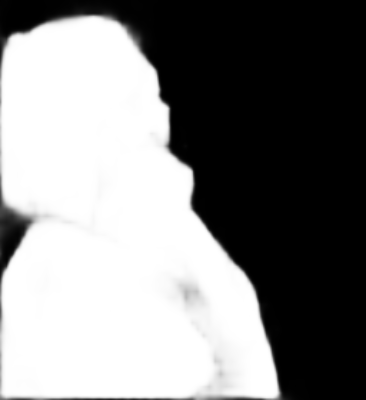}\vspace{2pt}
\includegraphics[width=1\linewidth]{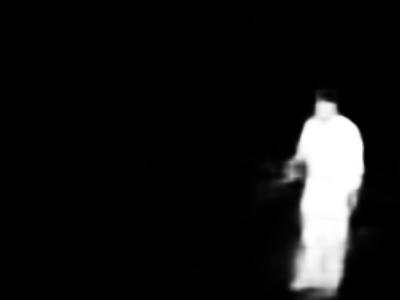}\vspace{2pt}
\end{minipage}}
\hspace{-1.2ex}
\subfigure[\tiny NLDF]{
\begin{minipage}[b]{0.087\linewidth}
\includegraphics[width=1\linewidth]{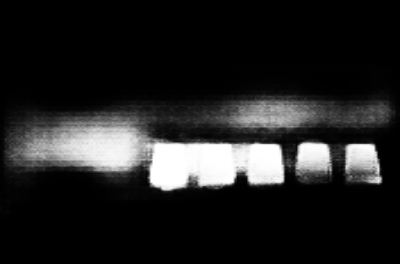}\vspace{2pt}
\includegraphics[width=1\linewidth]{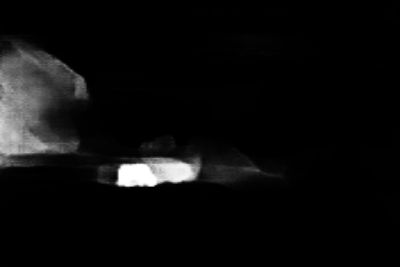}\vspace{2pt}
\includegraphics[width=1\linewidth]{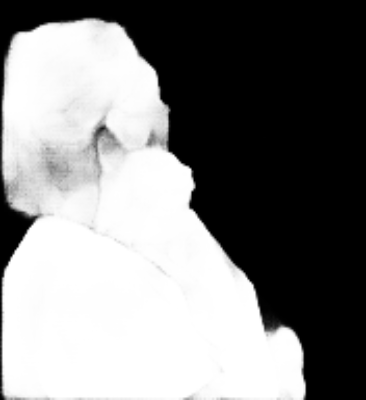}\vspace{2pt}
\includegraphics[width=1\linewidth]{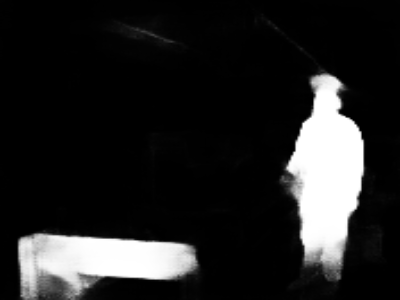}\vspace{2pt}
\end{minipage}}
\hspace{-1.2ex}
\subfigure[\tiny \textbf{\emph{NLDF-CPD-A}}]{
\begin{minipage}[b]{0.087\linewidth}
\includegraphics[width=1\linewidth]{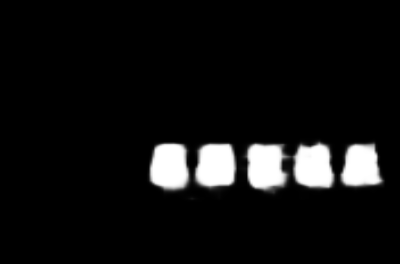}\vspace{2pt}
\includegraphics[width=1\linewidth]{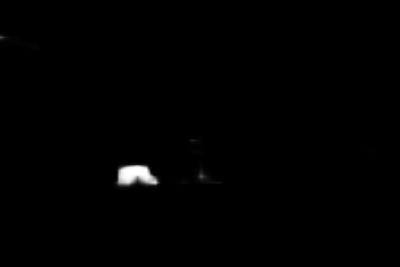}\vspace{2pt}
\includegraphics[width=1\linewidth]{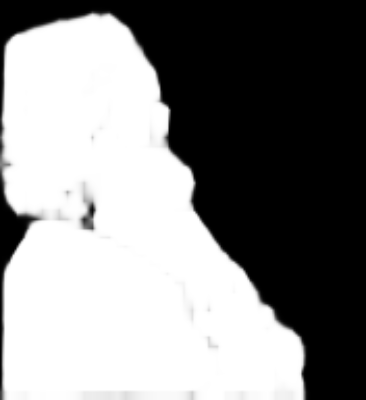}\vspace{2pt}
\includegraphics[width=1\linewidth]{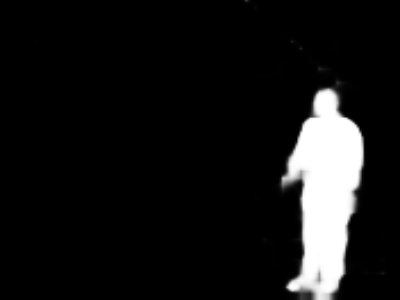}\vspace{2pt}
\end{minipage}}
\hspace{-1.2ex}
\subfigure[\tiny \textbf{\emph{NLDF-CPD}}]{
\begin{minipage}[b]{0.087\linewidth}
\includegraphics[width=1\linewidth]{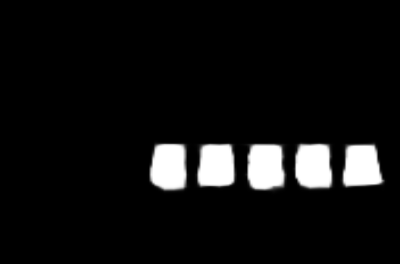}\vspace{2pt}
\includegraphics[width=1\linewidth]{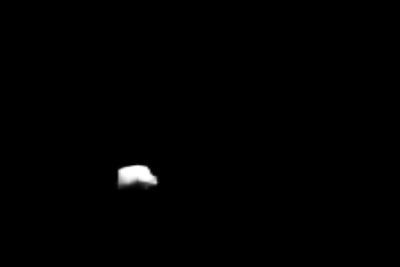}\vspace{2pt}
\includegraphics[width=1\linewidth]{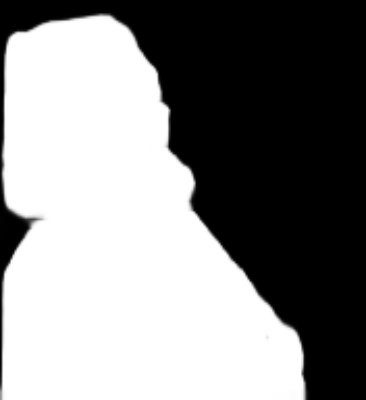}\vspace{2pt}
\includegraphics[width=1\linewidth]{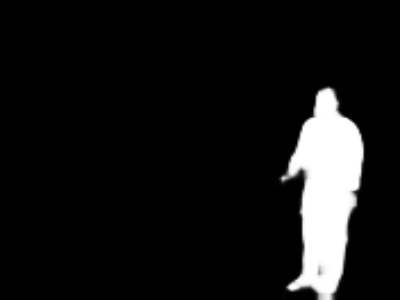}\vspace{2pt}
\end{minipage}}
\caption{\emph{Visual comparisons of original models (BMPM, Amulet, NLDF) with improved models (-CPD-A, -CPD).}}
\label{CPD_IN_OTHER_Figures}
\end{figure*}

\begin{table*}[htp]
  \centering
  \resizebox{\linewidth}{!}{
    \begin{tabular}{c|c|ccc|ccc|ccc|ccc|ccc}
    \hline
    \multicolumn{1}{c|}{\multirow{2}[0]{*}{Settings}} & \multicolumn{1}{c|}{\multirow{2}[0]{*}{FPS}} & \multicolumn{3}{c|}{ECSSD~\cite{Dataset-ECSSD}} & \multicolumn{3}{c|}{HKU-IS~\cite{2015MDF}} & \multicolumn{3}{c|}{DUT-OMRON~\cite{Dataset-DUT-OMRON}} & \multicolumn{3}{c|}{DUTS~\cite{Dataset-DUTS}} & \multicolumn{3}{c}{PASCAL-S~\cite{Dataset-PASCAL-S}} \\
    \cline{3-17}
          &       & maxF & avgF & MAE   & maxF & avgF & MAE & maxF & avgF & MAE & maxF & avgF & MAE & maxF & avgF & MAE \\
    \hline
    Conv2\_2 & 38 & \textbf{{\color{red}0.936}} & 0.903 & 0.042 & \textbf{{\color{red}0.925}} & 0.884 & 0.036 & 0.792 & 0.720 & 0.063 & 0.861 & 0.778 & 0.048 & 0.865 & 0.810 & 0.076 \\
    Conv3\_3 & 66 & \textbf{{\color{red}0.936}} & \textbf{{\color{red}0.915}} & \textbf{{\color{red}0.040}} & 0.924 & \textbf{{\color{red}0.896}} & \textbf{{\color{red}0.033}} & \textbf{{\color{red}0.794}} & \textbf{{\color{red}0.745}} & \textbf{{\color{red}0.057}} & \textbf{{\color{red}0.864}} & \textbf{{\color{red}0.813}} & \textbf{{\color{red}0.043}} & \textbf{{\color{red}0.866}} & \textbf{{\color{red}0.825}} & 0.074  \\
    Conv4\_3 & \textbf{{\color{red}90}} & 0.931 & 0.910 & 0.041 & 0.920 & 0.890 & 0.034 & 0.787 & 0.737 & 0.059 & 0.855 & 0.801 & 0.045 & 0.863 & 0.824 & \textbf{{\color{red}0.072}}\\
    Full Decoder & 30 & 0.922 & 0.891 & 0.051 & 0.911 & 0.873 & 0.042      & 0.758 & 0.692 & 0.070 & 0.843 & 0.766 & 0.050 & 0.853 & 0.807 & 0.077 \\
    \hline
    \end{tabular}}
  \caption{\emph{Comparison of the proposed model with different optimization layers and no optimization layer (full decoder).}}
  \label{OptimizationLayer}
\end{table*}

\noindent\textbf{Evaluation Metrics. } We adopt two metrics: mean absolute error (MAE) and F-measure (maxF). We adopt mean absolute error (MAE) and F-measure as our evaluation metrics. According to the different ways for saliency map binarization, there exist two ways to compute F-measure~\cite{2015SalObjBenchmark}. One is maximum F-measure (denoted as maxF), which is adopted in ~\cite{2018DSS,2018PiCANet,2017NLDF,2018BMPM}. The other is average F-measure (denoted as avgF), which is adopted in~\cite{2017SRM,2018DGRL,2017Amulet,2018PAGR}. For fairly comparison, we compute both maxF and avgF.



%
\noindent \textbf{Implementation Details.} We implement the proposed model based on the Pytorch\footnote{https://pytorch.org/} framework and a GTX $1080$Ti GPU is used for acceleration. Following previous works~\cite{2018PiCANet,2017SRM,2018DGRL,2018BMPM,2018PAGR}, we train the proposed model on the training set of DUTS~\cite{Dataset-DUTS} dataset. The parameters of the bifurcated backbone network are initialized by VGG$16$~\cite{2014VGG}. We initialize the other convolutional layers using the default setting of the Pytorch. All training and test images are resized to $352\times 352$. Any post-processing procedure (\emph{e.g.} CRF~\cite{2011CRF}) is not applied in this paper. The proposed model is trained by Adam optimizer~\cite{2014Adam}. The batch size is set as $10$ and the initial learning rate is set as $10^{-4}$ and decreased by $10\%$ when training loss reaches a flat. It takes nearly six hours for training the proposed model. The code is available at \url{https://github.com/wuzhe71/CPD}.

\subsubsection{Comparison with State-of-the-arts}
We compare the proposed model with eight state-of-the-art deep salient object detection algorithms, including NLDF~\cite{2017NLDF}, Amulet~\cite{2017Amulet}, DSS~\cite{2018DSS}, SRM~\cite{2017SRM}, BMPM~\cite{2018BMPM}, PAGR~\cite{2018PAGR}, DGRL~\cite{2018DGRL} and PiCANet~\cite{2018PiCANet}. We implement these models with available source codes or directly evaluate saliency maps provided by authors. Especially, NLDF, Amulet and DSS are originally trained on MSRA$10$K~\cite{2015GlobalContrast} dataset or MSRA-B~\cite{Dataset-MSRA-B} dataset (there is a large overlap between these two datasets). Hence we re-train these three models on DUTS dataset as other models for fairly comparison. We find that training on DUTS dataset will make deep models work better in complex scenes. Besides, we also train the proposed model on MSRA-B dataset to compare with these three original models, and the results are reported in supplementary material.

In Table.~\ref{All_Results}, we show the quantitative comparison results. Considering some works use ResNet$50$ as the backbone, we also train the proposed model on the basis of this backbone network.
ResNet$50$ contains four convolutional blocks, and we set the last layer of the second block as the optimization layer. Then we utilize the last two blocks to design the two branches. In Table.~\ref{All_Results}, the results of the attention branch (denoted as ``-A'') of the proposed model are also reported. Moreover, we compare the average execution time with the other models on DUTS dataset, and all scores are tested on our platform (PAGR only provides saliency maps). It is obvious that the proposed model outperforms all other models in most cases and it runs much faster than existing models. Only PiCANet-R obtains higher maxF score than the proposed model on DUT-OMRON dataset. However, our model runs about $12$ times faster than PiCANet-R. More specially, compared to the improvements on maxF and MAE, we obtain a larger improvement on avgF. This demonstrates that the proposed model works much better in uniformly highlighting salient objects. In addition, we can find that the results of our attention branch also achieve comparable results with other models. Meanwhile, the proposed model only with the attention branch runs faster. This indicates that the proposed model provides two-level saliency maps for real-time applications.

In Fig.~\ref{All_Figures}, we show the qualitative comparison on some challenging cases: small object, complex scenes, multiple objects and large object. Even though we discard the low-level features of backbone network, our model still recovers precise boundaries of salient objects, and the small object is still accurately segmented. Moreover, the proposed model segments more uniform salient objects than the compared models. It is consistency with the results in Table.~\ref{All_Results} that our model achieves more improvement in avgF score than MAE and maxF. This phenomenon is owing to the joint training strategy of the proposed model. On one hand, the supervised attention map of the attention branch makes the detection branch further concentrate on salient objects. On the other hand, when training the proposed model, the gradient of the detection branch also back propagates to the attention branch. This training mechanism gradually promotes the proposed model to focus on salient objects. More visual comparison results are shown in supplementary material.

\subsubsection{Application in Existing Models}
Through integrating features of each branch via using aggregation algorithms proposed in existing models, our framework can be utilized to improve these works. In this paper, we apply the proposed framework in three deep aggregation models (BMPM, Amulet, NLDF). NLDF adopts a typical U-Net architecture, BMPM proposes a bi-directional decoder with gate function and Amulet integrates multi-level feature maps in multiple resolutions. We implement the improved models in their respectively default deep learning library (tensorflow~\cite{Tensorflow} for BMPM and NLDF, caffe~\cite{Caffe} for Amulet). For BMPM and NLDF, we train the improved models (denoted as BMPM-CPD and NLDF-CPD) by using default settings, and it only needs to change the learning rate from the original $10^{-6}$ to $10^{-5}$. For Amulet, we train the improved model (denoted as Amulet-CPD) by using the completely same settings as the original model.

In Table.~\ref{CPD_IN_OtherModels}, we show the quantitative results of the original models and the improved models (-CPD-A, -CPD) on five benchmark datasets. We can see that each improved model outperform its original model. More specially, the improved models obtain a large improvement on the two most challenging DUT-OMRON and DUTS datasets. In addition, the improved models (-CPD and -CPD-A) runs about $2$ and $3$ times faster than the original models respectively. In conclusion, the proposed cascaded partial decoder framework can be used to improve deep aggregation models with different kinds of decoders. In Fig.~\ref{CPD_IN_OTHER_Figures}, we show the qualitative results on challenge cases: multiple objects, small object, large object and complex scene. The upper two rows show that the improved models further focus on target regions and suppress distractions. The under two rows show that the improved model further highlights the whole objects.

\begin{table}[t]
  \centering
  \resizebox{\linewidth}{!}{
    \begin{tabular}{c|ccc|ccc}
    \hline
    \multicolumn{1}{c|}{\multirow{2}[0]{*}{Settings}} & \multicolumn{3}{c|}{DUTS~\cite{Dataset-DUTS}} & \multicolumn{3}{c}{PASCAL-S~\cite{Dataset-PASCAL-S}} \\
    \cline{2-7}
          & maxF  & avgF & MAE   & maxF  & avgF & MAE \\
    \hline
    \textbf{\emph{CPD}} (with ia)  & 0.862 & 0.803 & 0.045 & 0.862 & 0.821 & 0.075 \\
    \textbf{\emph{CPD}} (with ha)  & \textbf{{\color{red}0.864}} & \textbf{{\color{red}0.813}} & \textbf{{\color{red}0.043}} & \textbf{{\color{red}0.866}} & \textbf{{\color{red}0.825}} & \textbf{{\color{red}0.074}} \\
    \hline
    \hline
    \textbf{\emph{Amulet-CPD}} (with ia)& 0.842 & 0.763 & 0.056 & 0.849 & 0.794 & 0.087 \\
    \textbf{\emph{Amulet-CPD}} (with ha)& \textbf{{\color{red}0.845}} & \textbf{{\color{red}0.771}} & \textbf{{\color{red}0.055}} & \textbf{{\color{red}0.851}} & \textbf{{\color{red}0.801}} & \textbf{{\color{red}0.085}} \\
    \hline
    \hline
    \textbf{\emph{BMPM-CPD}} (with ia) & 0.865 & 0.791 & 0.045 & 0.867 & 0.818 & \textbf{{\color{red}0.072}} \\
    \textbf{\emph{BMPM-CPD}} (with ha) & \textbf{{\color{red}0.870}} & \textbf{{\color{red}0.808}} & \textbf{{\color{red}0.044}} & \textbf{{\color{red}0.868}} & \textbf{{\color{red}0.822}} & \textbf{{\color{red}0.072}} \\
    \hline
    \hline
    \textbf{\emph{\textbf{NLDF-CPD}}} (with ia) & 0.838 & 0.777 & 0.051 & 0.840 & 0.793 & 0.084 \\
    \textbf{\emph{NLDF-CPD}} (with ha) & \textbf{{\color{red}0.842}} & \textbf{{\color{red}0.786}} & \textbf{{\color{red}0.048}} & \textbf{{\color{red}0.843}} & \textbf{{\color{red}0.800}} & \textbf{{\color{red}0.080}} \\
    \hline
    \end{tabular}}
  \caption{\emph{Comparison of initial attention (ia) and holistic attention (ha) in four models (the proposed model and three improved models).}}
  \label{SoftAttentoin}
\end{table}

\subsubsection{Analysis of the Proposed Framework}
\noindent\textbf{Effectiveness of holistic attention.} Here we demonstrate the effectiveness of the proposed holistic attention model in the proposed model and the three improved models. We compare these models with holistic attention and the models with initial attention, and the results are shown in Table.\ref{SoftAttentoin}. It is shown that holistic attention outperforms initial attention.
\noindent\textbf{Selection of Optimization Layer.} In the proposed model, we set Conv$3\_3$ layer as the optimization layer. Here we compared the proposed model with different optimization layers (Conv$2\_2$ and conv$4\_3$). In addition, we also report the results of no optimization layer, which means integrating all-level features via the proposed decoder. We do not test the proposed model with Conv$1\_2$ optimization layer because this setting will increase the computation cost via adding one more full decoder; thus requirements of reducing computation cost will not be achieved. The comparison results on five benchmark datasets are shown in Table.~\ref{OptimizationLayer}. In conclusion, we set the conv$3\_3$ layer as the optimization layer considering its best performance. When we refine the shallower feature (Conv$2\_2$), the computation complexity increases and the performance decreases. The reason might be that the feature of shallower layer has not been enough trained. When we refine the deep feature (Conv$4\_3$), the computation cost and the performance both decrease. This is because that resolution of the feature in the Conv$4\_3$ layer is smaller. The accuracy and efficiency of settings (Conv$2\_2$ and Conv$4\_3$) both outperform the full decoder, which validates the effectiveness of the proposed framework.

\begin{figure}[t]
\centering
\subfigure[Image]{
\begin{minipage}[b]{0.23\linewidth}
\includegraphics[width=1\linewidth]{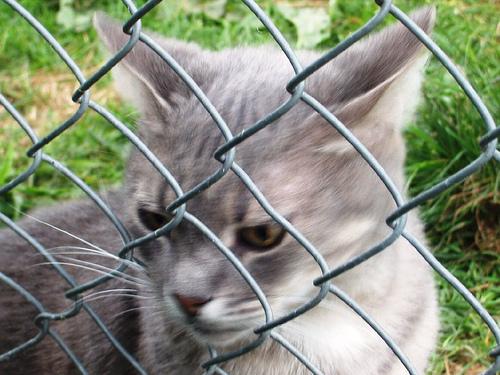}\vspace{2pt}
\includegraphics[width=1\linewidth]{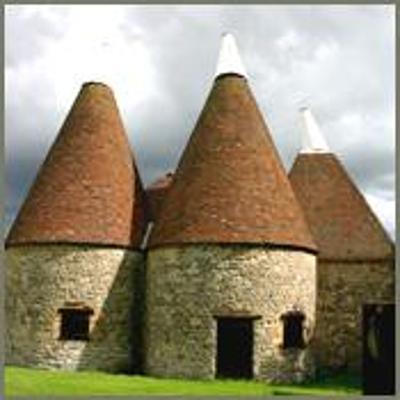}\vspace{2pt}
\end{minipage}}
\hspace{-1.2ex}
\subfigure[GT]{
\begin{minipage}[b]{0.23\linewidth}
\includegraphics[width=1\linewidth]{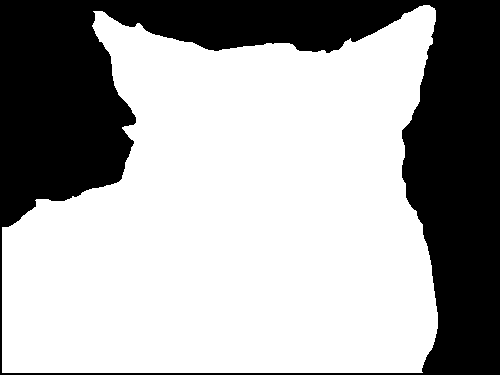}\vspace{2pt}
\includegraphics[width=1\linewidth]{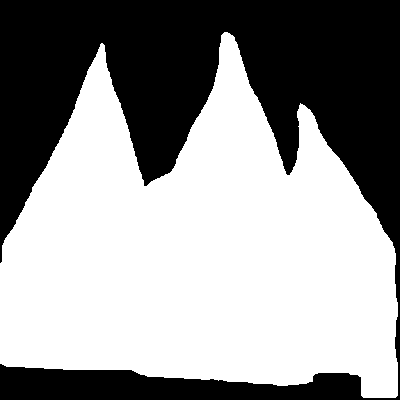}\vspace{2pt}
\end{minipage}}
\hspace{-1.2ex}
\subfigure[CPD-A]{
\begin{minipage}[b]{0.23\linewidth}
\includegraphics[width=1\linewidth]{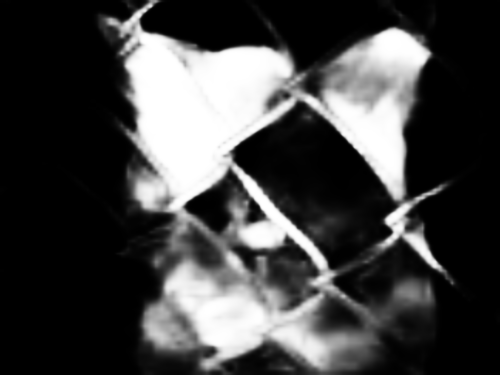}\vspace{2pt}
\includegraphics[width=1\linewidth]{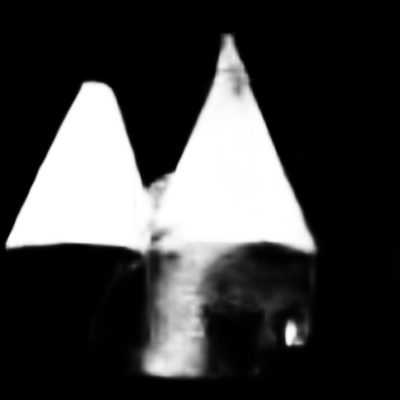}\vspace{2pt}
\end{minipage}}
\hspace{-1.2ex}
\subfigure[CPD]{
\begin{minipage}[b]{0.23\linewidth}
\includegraphics[width=1\linewidth]{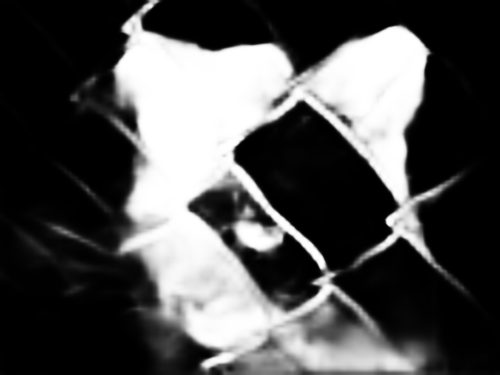}\vspace{2pt}
\includegraphics[width=1\linewidth]{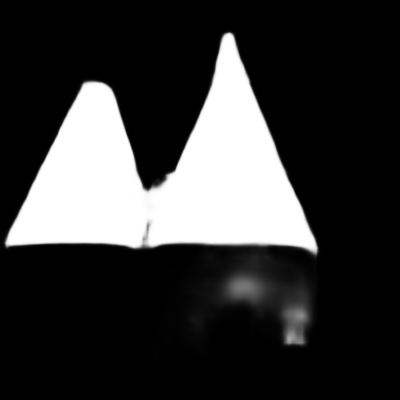}\vspace{2pt}
\end{minipage}}
\caption{\emph{Some Failure examples of the proposed model. When the attention branch only localizes a small part of target regions, our model performs poorly.}}\label{Failures}
\end{figure}

\noindent\textbf{Failure Examples.} The performance of the proposed model relies on the accuracy of the attention branch. When the attention branch detects clutters as target regions, our model will obtain wrong results. In Fig.~\ref{Failures}, we show some failure examples of our model. When a large target region is not segmented correctly, the proposed model is unable to segment the whole objects.

\subsection{Application in Other Tasks}
In this paper, we also evaluate the proposed model on other two binary segmentation tasks: shadow detection and portrait segmentation.

\noindent \textbf{Shadow Detection.} We re-train our model on the training set of SBU~\cite{Dataset-SBU-Shadow} dataset and test the model on three public shadow detection datasets: test set of SBU, ISTD~\cite{Dataset-ISTD} and UCF~\cite{Dataset-UCF-Shadow}. Moreover, we apply the widely-used metric BER (balanced error rate) for quantitative comparison. We compare our method with five deep shadow detection methods: JDR~\cite{Dataset-ISTD}, DSC~\cite{2018DSC}, DC-DSPF~\cite{2018DC-DSPF}, scGAN~\cite{2017scGAN}, StackedCNN~\cite{Dataset-SBU-Shadow}. In addition, we re-train three salient object detection models for shadow detection: NLDF~\cite{2017NLDF}, DSS~\cite{2018DSS}, BMPM~\cite{2018BMPM}. The results are shown in Table.~\ref{Shadow_Results}, and the proposed model outperforms the other models in all cases.

\begin{table}[t]
  \centering
  \resizebox{0.85\linewidth}{!}{
    \begin{tabular}{c|c|c|c}
    \hline
          & SBU~\cite{Dataset-SBU-Shadow}   & ISTD~\cite{Dataset-ISTD}  & UCF~\cite{Dataset-UCF-Shadow} \\
    \hline
    Method & BER$\downarrow$  & BER$\downarrow$   & BER$\downarrow$ \\
    \hline
    NLDF~\cite{2017NLDF}  & 7.02 & 7.50 & 7.69  \\
    \hline
    DSS~\cite{2018DSS}   & 7.00 & 10.48 & 10.56 \\
    \hline
    BMPM~\cite{2018BMPM}  & 6.17 & 7.10 & 8.09 \\
    \hline
    scGAN~\cite{2017scGAN} & 9.10 & 8.98 & 11.50 \\
    \hline
    StackedCNN~\cite{Dataset-SBU-Shadow} & 11.00 & 10.45 & 13.00 \\
    \hline
    JDR~\cite{Dataset-ISTD}   & 8.14 & 7.35 & 11.23  \\
    \hline
    DC-DSPF~\cite{2018DC-DSPF} & 4.90 & - & 7.90 \\
    \hline
    DSC~\cite{2018DSC}   & 5.59 & 8.24 & 8.10\\
    \hline
    \textbf{\emph{CPD}} (ours) & \textbf{{\color{red}4.19}}  & \textbf{{\color{red}6.76}}  & \textbf{{\color{red}7.21}} \\
    \hline
    \end{tabular}}
  \caption{\emph{Comparing the proposed method with state-of-the-arts for shadow detection (DSC, DC-DSPF, JDR, StackedCNN, scGAN), and for salient object detection (Amulet, NLDF, BMPM, DSS). }}
  \label{Shadow_Results}
\end{table}

\begin{table}[t]
  \centering
  \resizebox{\linewidth}{!}{
    \begin{tabular}{c|cccc|c}
    \hline
    Methods & \multicolumn{1}{c}{PFCN+~\cite{2016PortraiSeg}} &
    \multicolumn{1}{c}{NLDF~\cite{2017NLDF}} & \multicolumn{1}{c}{DSS~\cite{2018DSS}} & \multicolumn{1}{c|}{BMPM~\cite{2017Amulet}} & \multicolumn{1}{c}{\textbf{\emph{CPD}} (Ours)} \\
    \hline
    Mean IoU & 95.90\% & 95.60\% & 96.20\% & 96.20\% & \textbf{{\color{red}96.60\%}} \\
    \hline
    \end{tabular}}
  \caption{\emph{Quantitative Comparison on Portrait Segmentation.}}
  \label{PortraitSeg}
\end{table}

\noindent \textbf{Portrait Segmentation.} We use the data from~\cite{2016PortraiSeg}. And we re-train NLDF, DSS, BMPM on this dataset. The results are shown in Table~\ref{PortraitSeg}. It can be seen that the proposed model outperforms existing algorithms.

\section{Conclusion}
In this paper, we propose a novel cascaded partial decoder framework for fast and accurate salient object detection. When constructing decoders, the proposed framework discards features of shallower layers to improve the computational efficiency, and utilizes generated saliency map to refine features to improve the accuracy. We also propose a holistic attention module to further segment the whole salient objects and an effective decoder to abstract discriminative features and quickly integrate multi-level features. The experiments show that our model achieves state-of-the-art performance on five benchmark datasets and runs much faster than existing deep models. To prove the generalization of the proposed framework, we apply it to improve existing deep aggregation models and significantly improve their accuracy and efficiency. Besides, we validate the effectiveness of the proposed model in two tasks of shadow detection and portrait segmentation.

\noindent \textbf{Acknowledgement.} This work was supported by the University of Chinese Academy of Sciences, in part of National $61472389$, $61620106009$, $61772494$, U$1636214$, $61771457$, $61732007$, in part by Key Research Program of Frontier Sciences, CAS: QYZDJ-SSW-SYS$013$.

{\small
\bibliographystyle{ieee}
\bibliography{egbib}
}

\end{document}